\documentclass{article}

\PassOptionsToPackage{dvipsnames}{xcolor}

\usepackage[main,final]{neurips_2026}

\usepackage[utf8]{inputenc}
\usepackage[T1]{fontenc}
\usepackage[colorlinks=true,citecolor=blue,linkcolor=red,urlcolor=blue]{hyperref}
\usepackage{url}
\usepackage{microtype}
\usepackage{graphicx}
\usepackage{subcaption}
\usepackage{booktabs}
\IfFileExists{svg.sty}{\usepackage{svg}}{}
\usepackage{algorithm}
\usepackage{algorithmic}
\usepackage{amsmath}
\usepackage{amssymb}
\usepackage{amsfonts}
\usepackage{mathtools}
\usepackage{amsthm}
\usepackage{bm}
\usepackage{bbm}
\usepackage{enumitem}
\usepackage{float}
\usepackage{wrapfig}
\usepackage{placeins}
\usepackage{mdframed}
\usepackage[nameinlink]{cleveref}
\usepackage{tikz}
\usepackage{pgfplots}
\usepackage[acronym]{glossaries}

\pgfplotsset{compat=1.18}

\usetikzlibrary{arrows.meta,positioning,calc,fit,decorations.pathreplacing}

\numberwithin{equation}{section}

\newcommand{\Unif}{\mathcal{U}}

\glsdisablehyper
\newacronym{fm}{FM}{flow matching}
\newacronym{dm}{DM}{diffusion model}
\newacronym{dfm}{DFM}{discrete flow model}
\newacronym{vfm}{VFM}{Variational Flow Matching}
\newacronym{gem}{GEM}{Graph Energy Matching}
\newacronym{gw}{GW}{Gromov–Wasserstein}
\newacronym{fgw}{FGW}{Fused Gromov–Wasserstein}
\newacronym{gwot}{GWOT}{Gromov–Wasserstein optimal transport}
\newacronym{mh}{MH}{Metropolis–Hastings}
\newacronym{ebm}{EBM}{energy-based model}
\newacronym{ot}{OT}{optimal transport}
\newacronym{mcmc}{MCMC}{Markov chain Monte Carlo}
\newacronym{ctmc}{CTMC}{continuous-time Markov chain}
\newacronym{ode}{ODE}{ordinary differential equation}
\newacronym{sde}{SDE}{stochastic differential equation}
\newacronym{jko}{JKO}{Jordan–Kinderlehrer–Otto}
\newacronym{cfg}{CFG}{classifier-free guidance}
\newacronym{gnn}{GNN}{graph neural network}
\newacronym{mlp}{MLP}{multilayer perceptron}
\newacronym{mae}{MAE}{mean absolute error}
\newacronym{qed}{QED}{quantitative estimate of drug-likeness}
\newacronym{tpsa}{TPSA}{topological polar surface area}
\newacronym{vu}{V.U.}{valid-unique}
\newacronym{vun}{V.U.N.}{valid-unique-novel}
\newacronym{cvun}{C.V.U.N.}{conditional valid-unique-novel}
\newacronym{cvu}{C.V.U.}{conditional valid-unique}
\newacronym{fcd}{FCD}{Fr{\'e}chet ChemNet Distance}

\definecolor{cb_blue}{HTML}{4477AA}
\definecolor{cb_orange}{HTML}{EE7733}
\definecolor{cb_red}{HTML}{CC3311}
\definecolor{cb_teal}{HTML}{009988}
\definecolor{cb_gray}{HTML}{BBBBBB}
\definecolor{cb_green}{HTML}{228833}
\definecolor{cb_lightgray}{HTML}{AAAAAA}

\newcommand{\xdata}{x^{\text{data}}}
\newcommand{\xzero}{x^{0}}

\newcommand{\pidata}{\pi_{\text{data}}}
\newcommand{\pizero}{\pi_{0}}

\newcommand{\Vtheta}{V_\theta}
\newcommand{\qgreedy}{q_{\text{greedy}}}
\newcommand{\qmixing}{q_{\text{mixing}}}

\title{Graph Energy Matching: Transport-Aligned Energy-Based Modeling for Graph Generation}

\author{%
  Michal Balcerak\textsuperscript{1,2,}\setcounter{footnote}{1}%
  \thanks{Much of this work was done during the author's previous appointment at the University of Zurich.}\setcounter{footnote}{0} \\
  {\normalsize\texttt{michal\_balcerak@harvard.edu}}
  \And
  Suprosanna Shit\textsuperscript{3}
  \And
  Chinmay Prabhakar\textsuperscript{3}
  \AND
  Sebastian Kaltenbach\textsuperscript{1}
  \And
  Michael S. Albergo\textsuperscript{1,2}
  \And
  Yilun Du\textsuperscript{1,2,}\thanks{Contributed equally as senior authors.}
  \And
  Bjoern Menze\textsuperscript{3,}\footnotemark[1]
  \AND
  \normalfont
  \textsuperscript{1}Harvard University\quad
  \textsuperscript{2}Kempner Institute\\[0.3ex]
  \textsuperscript{3}University of Zurich
}

\begin{document}

\maketitle

\begin{abstract}
Generative modeling of discrete data, such as graphs, underpins many scientific and industrial applications, including molecular discovery and materials design. In these domains, probabilistic inference is particularly valuable, as it enables composable generation and principled incorporation of desired constraints, such as structural or functional properties. Energy-based models naturally support this goal by capturing relative likelihoods and enabling composable inference by directly enforcing constraints during inference. However, discrete energy-based models typically struggle with efficient and high-quality sampling, as off-support regions often contain spurious local minima, trapping samplers and causing training instabilities, resulting in a fidelity gap compared to discrete diffusion models. To address this gap, we introduce \emph{Graph Energy Matching (GEM)}, a discrete generative framework inspired by the Jordan--Kinderlehrer--Otto (JKO) transport-map optimization perspective. GEM learns a permutation-invariant potential energy that simultaneously guides discrete transport from noise toward high-likelihood graph regions and refines samples within these regions. We further introduce a sampling protocol leveraging an energy-based switching strategy, seamlessly bridging rapid, gradient-guided transport and a local mixing regime for effective exploration. On molecular graph benchmarks, GEM matches or surpasses strong discrete diffusion baselines on most reported metrics. Beyond improving generation quality, GEM's relative likelihood modeling enables targeted exploration, facilitating compositional generation, property-constrained sampling, and interpolation between graphs. Project page: \url{https://michalbalcerak.ai/graph-energy-matching/}.
\end{abstract}

\section{Introduction}
\label{sec:introduction}

Generative modeling of discrete objects, such as graphs, 
is a central challenge across many scientific and industrial applications, 
including drug discovery and materials design \citep{jin2018junction,stokes2020deep,xie2022crystal}.  While domain-driven simulations explicitly encode physical knowledge \citep{noe2020machine}, 
they quickly become intractable due to the combinatorial complexity of the underlying data distribution \citep{polishchuk2013estimation}, 
highlighting the need for generative models that efficiently capture implicit, domain-specific patterns.

Current state-of-the-art models for unconditional graph generation predominantly rely on discrete diffusion-like processes, implemented either as discrete-time denoising steps~\citep{vignac2022digress} or \glspl{ctmc}~\citep{campbell2024generative,siraudin2024cometh,qinmadeira2024defog}. These methods generate graphs by progressively denoising samples from a simple prior distribution toward the target data distribution. For instance, DeFoG~\citep{qinmadeira2024defog} learns \gls{ctmc} transition rates to remove noise from corrupted graph states, while \gls{vfm}~\citep{eijkelboom2024vfm, eijkelboom2025controlledvfm} parameterizes stochastic trajectories that continuously transform samples from noise distributions into clean graphs. Although these approaches yield high-quality unconditional samples, they define distributions implicitly via noisy intermediate states, lacking an explicit model directly on clean graphs. This implicit representation complicates the enforcement of user-specified properties or constraints, as guidance must operate on intermediate, often off-manifold states, where predictors and conditioning signals become less reliable~\citep{vignac2022digress}.

\Glspl{ebm}~\citep{hinton2002training, lecun2006tutorial, du2019implicit} offer a complementary perspective that naturally addresses these limitations. \Glspl{ebm} represent relative probability structure through a scalar energy function, enabling direct incorporation of constraints, priors, or property-based objectives at inference time without retraining~\citep{du2025learning}. For graph generation, \glspl{ebm} offer particular advantages, as graph-structured constraints and domain-specific priors can be naturally integrated into the energy function. Historically, however, \glspl{ebm} have struggled with sample quality, primarily due to poor sampling efficiency in discrete, combinatorial domains, as exemplified by existing methods such as GraphEBM~\citep{liu2021graphebm}. Consequently, discrete \glspl{ebm} have predominantly served as scoring models for out-of-distribution detection rather than competitive generation~\citep{wu2023energy,fuchsgruber2024energy}.

Energy Matching~\citep{balcerak2025energy} bridges these perspectives by building on the \gls{jko} formulation~\citep{jordan1998variational} and its first-order optimality perspective~\citep{terpin2024learning,lanzetti2024variational,lanzetti2025firstorder}, coupling transport toward the data distribution with refinement through a time-indexed sequence of sampling steps. In these continuous formulations, dynamics typically follow \glspl{ode} or \glspl{sde}, guided by a learned scalar potential that defines both the transport direction and refinement by capturing a Boltzmann-like probability structure of the data. 

However, Energy Matching formulations operate in continuous spaces and rely on time-indexed dynamics defined over ambient continuous domains. This approach does not naturally extend to discrete spaces, where generation fundamentally requires discrete sampling characterized by abrupt, local state transitions rather than continuous trajectories. This highlights the need for a new framework enabling high-quality graph generation while explicitly capturing the relative likelihood structure, thus recovering the compositional and constraint-handling advantages of \glspl{ebm}.

\begin{mdframed}[hidealllines=true,backgroundcolor=BurntOrange!8]
\paragraph{Contributions.}
We introduce \glsadd{gem}\emph{\gls{gem}}, a novel discrete energy-based generative model for graphs, achieving molecular graph generation quality that matches or surpasses leading discrete diffusion approaches on most reported metrics.  Beyond unconditional sampling, the learned relative-likelihood structure enables \gls{gem} to incorporate compositional constraints at inference time and facilitates the computation of geodesics between graphs.

We leverage a \gls{jko}-style transport map optimization perspective to define (i) a transport-aligned discrete proposal to rapidly move towards high-probability graphs, and (ii) a discrete mixing proposal that efficiently explores the learned data geometry through local graph edits. \gls{gem} provides a principled pathway toward solving limitations of energy-based discrete graph generation, paving the way for their broader applicability.
\end{mdframed}

\vspace{-0.25em}
\section{Preliminaries: Energy Matching and JKO }

Energy matching learns a scalar potential \( V_\theta:\mathbb{X}
\to\mathbb{R} \), parameterized by \(\theta\), on a continuous data space \(\mathbb{X}\). It simultaneously serves two complementary roles: (i)~off-manifold guidance: directing samples toward the data manifold via the gradient field \(-\nabla_x V_\theta\), and (ii)~near-equilibrium refinement: characterizing the data distribution as an \gls{ebm}.

These objectives decompose model learning into a \emph{global} transport-alignment term and a \emph{local} density-refinement term. A time-dependent temperature schedule (e.g., $\epsilon(t)=0$ for $0 \leq t < 1$ for transport and $\epsilon(t)=\epsilon$ for $t \geq 1$ for mixing, promoting exploration) ensures that samples converge to the Gibbs measure $\rho_{\theta}(x) \propto \exp(-\frac{1}{\epsilon} V_\theta(x))$ as $t \to \infty$.

Energy Matching can be viewed through an optimization-in-probability-spaces perspective, expressed via the \gls{jko} scheme from density $\rho_t$ to $\rho_{t+\Delta t}$:
\begin{equation}
\label{jko_em}
\rho_{t+\Delta t}
=
\arg\min_{\rho}
\frac{1}{2\Delta t}\inf_{\gamma\in\Gamma(\rho_t,\rho)}\mathbb{E}_{(x,y)\sim\gamma}[c(x,y)]
+\int V_\theta(y)\,\rho(y)\,\mathrm{d}y
+\epsilon(t)\int \rho(y)\log\rho(y)\,\mathrm{d}y,
\end{equation}
where the cost \( c(x,y) \) quantifies displacement from \( x \) to \( y \), typically set as \( c(x,y)=\|x - y\|^2 \) in continuous spaces.

The coupling \(\gamma\in\Gamma(\rho_t,\rho)\) is a joint distribution that characterizes a transport plan. Intuitively, \(\gamma(x,y)\) encodes ``how much mass from each location \(x\sim\rho_t\) is transported to each new location \(y\sim\rho\)'', thus determining the optimal reassignment of mass that minimizes the expected transportation cost \(\mathbb{E}_{(x,y)\sim\gamma}[\,c(x,y)\,]\). In practice, couplings $\gamma$ can be precomputed or estimated during training, and $V_\theta$ is optimized accordingly, aligning it with the induced transport trajectories.

\paragraph{Sampling (two regimes).}
Sampling starts from a transport-aligned drift and switches to refinement via a prescribed time-dependent temperature
$\epsilon(t)$. In continuous time, this corresponds to the time-inhomogeneous \gls{sde}:
\begin{equation}
\label{em_sde}
\mathrm{d}x_t
=
-\nabla_{x_t} V_\theta(x_t)\,\mathrm{d}t
+\sqrt{2\,\epsilon(t)}\,\mathrm{d}W_t,
\end{equation}
which reduces to an \gls{ode} when $\epsilon(t)=0$ and becomes stochastic as $\epsilon(t)> 0$.
In practice, sampling corresponds to explicitly integrating \eqref{em_sde} over a fixed time interval.

\section{Graph Energy Matching}
\label{sec:gem}

\begin{figure*}[!t]
\centering

\definecolor{otColor}{RGB}{230,159,0}   
\definecolor{mhColor}{RGB}{0,114,178}   
\tikzset{state/.style={circle,draw,minimum size=11mm,inner sep=0pt}}

\begin{minipage}{\textwidth}
\centering
\begin{tikzpicture}[>=Stealth, every node/.style={font=\small}, node distance=1.2cm, remember picture]
\def\transportshift{-0.35cm}
\def\dividercomp{0.15cm}

\begin{scope}[xshift=\transportshift]
\node[state,fill=orange!20]                    (pi0b) {$\pi_{0}$};
\node[state,fill=orange!20,right=of pi0b]      (pi1b) {$\pi_{1}$};
\node[right=of pi1b]                           (dots1b) {$\dots$};
\node[state,fill=orange!20,right=of dots1b]    (pikb) {$\pi_{k}$};
\coordinate (transport-center) at ($(pi0b)!0.5!(pikb)$);

\draw[->,very thick,draw=otColor,shorten <=7.35pt,shorten >=7.35pt] (pi0b) -- (pi1b);
\draw[->,very thick,draw=otColor,shorten <=7.35pt,shorten >=7.35pt] (pi1b) -- (dots1b);
\draw[->,very thick,draw=otColor,shorten <=7.35pt,shorten >=7.35pt] (dots1b) -- (pikb);

\node[below=16pt of transport-center,align=center,font=\scriptsize]
      {greedy discrete proposals on samples\\(direction $-\nabla_x V_{\theta}(x)$ $\leftrightarrow$ transport map direction)};
\end{scope}

\node[coordinate] (dividerx) at ($(pikb.east)+(0.10,0) + (\dividercomp,0)$) {};
\coordinate (divider-top) at ($(dividerx)+(0,0.95)$);
\coordinate (divider-bottom) at ($(dividerx)+(0,-0.95)$);

\node[right=1.0cm of dividerx]                    (dots2b) {$\dots$};
\node[state,fill=blue!15,right=0.8cm of dots2b]  (pidatab) {$\pi_{\text{data}}$};
\node[right=2mm of pidatab]                      (eqb) {$=$};
\node[state,fill=blue!15,right=2mm of eqb]       (pistb) {$\pi_{\theta}$};

\coordinate (mh-start) at ($(dividerx)+(0.05cm,0)$);
\draw[->,very thick,draw=mhColor,shorten <=7pt,shorten >=7pt] (mh-start) -- (dots2b);
\draw[->,very thick,draw=mhColor,shorten <=7pt,shorten >=7pt] (dots2b) -- (pidatab);

\coordinate (mixing-center) at ($(dots2b)!0.5!(pistb)$);
\node[below=16pt of mixing-center,align=center,font=\scriptsize]
      {multi-site, local, geometry-aware\\$\nabla_x V_\theta(x)$-informed jumps\\+ \glsentryshort{mh} acceptance};

\coordinate (phase-label-y) at ($(pi0b.north)+(0,0.70cm)$);
\node[align=center,anchor=south] at (phase-label-y -| transport-center)
      {\textbf{Transport Phase}\\[-1pt]\scriptsize samples far from data manifold: $V_{\theta}(x) \gg V_{\text{data}}^{\text{mean}}$};
\node[align=center,anchor=south] at (phase-label-y -| pidatab)
      {\textbf{Mixing Phase}\\[-1pt]\scriptsize samples near data manifold: $V_{\theta}(x) \sim V_{\text{data}}^{\text{mean}}$};

\node[anchor=north west,font=\normalsize]
  at ($(current bounding box.west |- phase-label-y)+(0,-1.5mm)$)
  {\textbf{Probability Distributions Perspective:}};
\draw[->,thick,draw=mhColor]
  ($(pistb.east)+(1pt,3pt)$) arc[start angle=160,end angle=-160,radius=6pt];

\end{tikzpicture}
\end{minipage}

\vspace{-0.04cm}

\begin{minipage}{\textwidth}
\centering
\makebox[\textwidth][l]{\raisebox{0.15cm}{\normalsize\textbf{Samples Perspective:}}}\\[1pt]
\begingroup
\setlength{\tabcolsep}{1.27pt}
\renewcommand{\arraystretch}{0.82}
\newcommand{\samplewidth}{0.104\textwidth}
\newcommand{\sampleshift}{1.5em}
\newcommand{\samplerightshift}{1.5em}
\newcommand{\sampleplaceholder}{\makebox[\samplewidth][c]{}}
\makebox[0pt][l]{\hspace*{-\sampleshift}\begin{tabular}[t]{ccccc}
\includegraphics[width=\samplewidth]{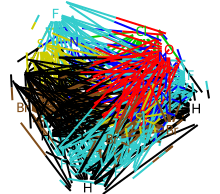} &
\includegraphics[width=\samplewidth]{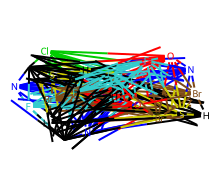} &
\includegraphics[width=\samplewidth]{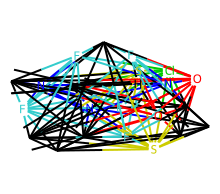} &
\includegraphics[width=\samplewidth]{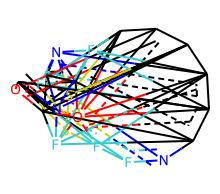} &
\includegraphics[width=\samplewidth]{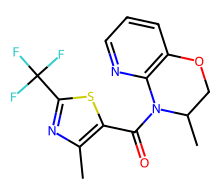} \\[-0.5em]
\includegraphics[width=\samplewidth]{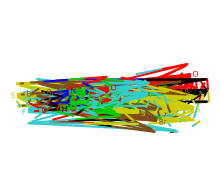} &
\includegraphics[width=\samplewidth]{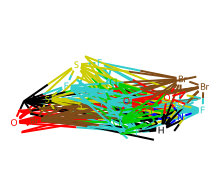} &
\includegraphics[width=\samplewidth]{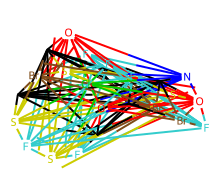} &
\includegraphics[width=\samplewidth]{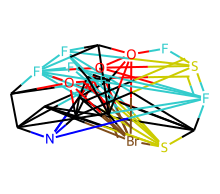} &
\includegraphics[width=\samplewidth]{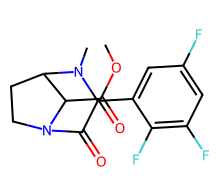} \\
\end{tabular}}%
\hspace*{\samplerightshift}\begin{tabular}[t]{cccccccc}
\sampleplaceholder & \sampleplaceholder & \sampleplaceholder & \sampleplaceholder & \sampleplaceholder &
\includegraphics[width=\samplewidth]{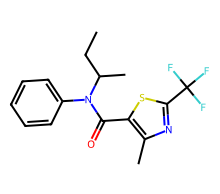} &
\includegraphics[width=\samplewidth]{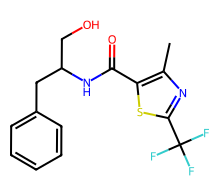} &
\includegraphics[width=\samplewidth]{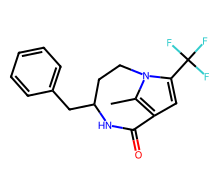} \\[-0.5em]
\sampleplaceholder & \sampleplaceholder & \sampleplaceholder & \sampleplaceholder & \sampleplaceholder &
\includegraphics[width=\samplewidth]{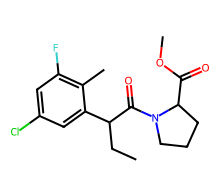} &
\includegraphics[width=\samplewidth]{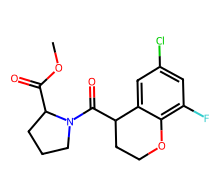} &
\includegraphics[width=\samplewidth]{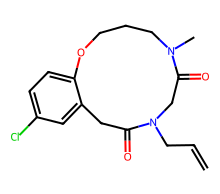} \\
\end{tabular}
\endgroup
\end{minipage}
\tikz[remember picture, overlay]\node (samples-bottom) {};
\begin{tikzpicture}[remember picture, overlay]
\draw[densely dashed,line width=1.2pt]
  ($(divider-top)+(0,8mm)$) -- ($(divider-top)!0.95!(divider-top |- samples-bottom)$);
\end{tikzpicture}

\caption{\textbf{\glsentryshort{gem} Sampling Overview.} Two perspectives on \glsentryshort{gem} sampling: a probability-distribution
view (top) of the two-phase \glsentryshort{mcmc} process, and a samples view (bottom) showing molecular trajectories
from MOSES. Sampling alternates between a transport phase, where gradient-informed,
greedy proposals rapidly move samples toward regions of high probability, and a mixing phase
employing \glsentryshort{mh} acceptance to ensure correct stationary distribution and efficient mixing between
modes. \textbf{Color key:} transport (\textcolor{otColor}{orange}), \glsentryshort{mh}
(\textcolor{mhColor}{blue}).}

\label{fig:sampling-regimes}
\end{figure*}

We extend the continuous Energy Matching formulation to discrete graph spaces by introducing two novel discrete proposals: one enabling transport-aligned rapid movements across graph spaces, and the other facilitating mixing within high-probability regions. Our formulation employs an energy-based switching mechanism that clearly separates \emph{transport} and \emph{mixing} sampling regimes without requiring explicit indexing of time or noise levels. Sampling is efficiently facilitated through deterministic greedy proposals in the transport regime and stochastic gradient-informed proposals during mixing. The sampler uses a temperature $\epsilon$ to control stochasticity: the transport regime is the deterministic $\epsilon\to 0$ limit, while the mixing regime uses $\epsilon>0$. Further, we propose a temperature-annealing strategy to efficiently generate novel samples from given data samples.

\paragraph{Graph representation.} 
We consider a discrete state space $\mathcal{X}:=\big(\mathcal{X}_{\mathrm{node}}\big)^{n}\times\big(\mathcal{X}_{\mathrm{edge}}\big)^{\binom{n}{2}}$ to represent undirected graphs. A graph $x\in \mathcal{X}$ contains a set of $n$ nodes $\mathcal{V}=\{1,\dots,n\}$ and edges $\mathcal{E}$ defining the connectivity between these nodes. Each node $i\in\mathcal{V}$ has a node class $x_i \in \mathcal{X}_{\mathrm{node}}$ (e.g.\ an atom type), and each edge $(i,j)\in\mathcal{E}$ has an edge class $x_{ij}\in\mathcal{X}_{\mathrm{edge}}$ (e.g.\ a bond type). The edge space $\mathcal{X}_{\mathrm{edge}}$ includes an element indicating the absence of an edge. The graph $x:=\big(x_\mathcal{V},x_\mathcal{E}\big)$ consists of $x_\mathcal{V}$ and $x_\mathcal{E}$ denoting categorical node and edge features, respectively, and are defined as:
\begin{equation}
x_\mathcal{V}  :=[x_i]_{i=1}^n\in(\mathcal{X}_{\mathrm{node}})^n;\quad
x_\mathcal{E}  :=[x_{ij}]_{ i<j}\in\big(\mathcal{X}_{\mathrm{edge}}\big)^{\binom{n}{2}}
\end{equation}

We embed the node and edge categorical feature vectors $x_\mathcal{V}$ and $x_\mathcal{E}$ using a real-valued one-hot encoding and treat them as continuous features via:
\[
\phi_{\mathcal{V}} :\mathcal{X}_{\mathrm{node}}\to\mathbb{R}^{l_{\mathrm{node}}};\quad
\phi_{\mathcal{E}} :\mathcal{X}_{\mathrm{edge}}\to\mathbb{R}^{l_{\mathrm{edge}}}
\]
where $l_{\mathrm{node}}=|\mathcal{X}_{\mathrm{node}}|$ and $l_{\mathrm{edge}}=|\mathcal{X}_{\mathrm{edge}}|$ denote the numbers of node and edge classes, respectively. We denote the real-valued node features $\hat{x}_{\mathcal{V}}\in\mathbb{R}^{n l_{\mathrm{node}}}$ and edge features $\hat{x}_{\mathcal{E}}\in\mathbb{R}^{\binom{n}{2}l_{\mathrm{edge}}}$, and the corresponding graph representation 
$\hat{x}\in\mathbb{R}^d$, where $d=n l_{\mathrm{node}}+\binom{n}{2}l_{\mathrm{edge}}$, as:
\begin{equation}
\hat{x}_\mathcal{V}:=\oplus_{i=1}^n \phi_\mathcal{V}(x_i);\quad
\hat{x}_\mathcal{E}:=\oplus_{i<j}\phi_\mathcal{E}(x_{ij});\quad
\hat{x}:=\hat{x}_\mathcal{V}\oplus \hat{x}_\mathcal{E}.
\end{equation}
where $\oplus$ denotes the column-wise concatenation operation.

\paragraph{Local and Permutation-Invariant Cost.} We measure local discrepancies between two graphs $x$ and $y$ using the embedding-space cost $
    c(x,y) := \tilde c(\hat{x},\hat{y})$,
which is differentiable in both arguments. The potential $V_\theta$ is permutation-invariant by construction (\Cref{app:perm-invariance}). %
A permutation $\sigma\in S_n$ acts on node indices and induces a relabeling of the graph: $(\sigma\cdot x)_i=x_{\sigma(i)}$ and $(\sigma\cdot x)_{ij}=x_{\sigma(i)\sigma(j)}$. To obtain a permutation-invariant matching, we first define a local cost ($c_{\mathrm{loc}}$) as:
\begin{equation}\label{eq:loc-cost}
c_{\mathrm{loc}}(x,y)
\;=\;
\lambda_\mathcal{V}\|\hat{x}_{\mathcal{V}}-\hat{y}_{\mathcal{V}}\|^2_2
+
\lambda_\mathcal{E}\|\hat{x}_{\mathcal{E}}-\hat{y}_{\mathcal{E}}\|^2_2,
\end{equation}
with $\lambda_\mathcal{V},\lambda_\mathcal{E}>0$ scaling node and edge contributions. Optimizing over node relabelings lifts the local cost to the hard-permutation \gls{fgw} cost \citep{vayer2019optimal}:
\begin{equation}\label{eq:fgw-cost}
c_{\mathrm{FGW}}(x,y)
=
\min_{\sigma\in S_n}
\left\{
c_{\mathrm{loc}}(x,\sigma\cdot y)
\right\}.
\end{equation}
In practice, we predominantly use the local cost $c_{\mathrm{loc}}$, as it provides a valid and efficient measure for scoring pairs of neighboring graphs, e.g.\ when evaluating displacements induced by local jump proposals. When permutation-invariant comparisons between non-local graph pairs are required, we instead employ the $c_{\mathrm{FGW}}$ cost. Exact computation is generally intractable, as it is linked to the Quadratic Assignment Problem (QAP) \citep{vayer2019optimal}, so we use fast approximations: histogram matching for noisy--clean pairs and node matching for molecule--molecule pairs (\Cref{app:matching}).

\begin{wrapfigure}[13]{r}{0.50\textwidth}
\vspace{-1.4em}
\centering
\resizebox{0.94\linewidth}{!}{%
\begin{tikzpicture}[>=Latex, font=\small]

\def\W{11.2}
\def\H{4.9}
\node[font=\small\bfseries] at (\W/2-0.7,\H+0.45)
{Discrete, geometry-aware energy-based proposal (local edits)};

\coordinate (xhat) at (3.8,1.55);

\pgfmathsetmacro{\rOne}{0.85}
\pgfmathsetmacro{\rTwo}{1.35}
\pgfmathsetmacro{\rThr}{1.85}

\draw[cb_lightgray, dashed, line width=0.7pt] (xhat) circle[radius=\rOne];
\draw[cb_lightgray, dashed, line width=0.7pt] (xhat) circle[radius=\rTwo];
\draw[cb_lightgray, dashed, line width=0.7pt] (xhat) circle[radius=\rThr];

\node[cb_lightgray, font=\scriptsize, align=center] at ($(xhat)+(-2.3,1.2)$)
{distance penalty\\isolines};

\node[cb_lightgray, font=\scriptsize\bfseries] at ($(xhat)+(115:\rOne)+(0.06,-0.24)$) {1 edit};
\node[cb_lightgray, font=\scriptsize\bfseries] at ($(xhat)+(115:\rTwo)+(0.06,-0.26)$) {2 edits};
\node[cb_lightgray, font=\scriptsize\bfseries] at ($(xhat)+(115:\rThr)+(0.06,-0.28)$) {3 edits};

\filldraw[black, fill=cb_red] (xhat) circle (2.2pt);
\node[font=\scriptsize, anchor=north east] at ($(xhat)+(-0.06,-0.12)$) {$\hat{x}$};
\node[cb_lightgray, font=\scriptsize\bfseries, anchor=west] at ($(xhat)+(0.12,-0.02)$) {stay};

\pgfmathsetmacro{\gdeg}{30} 
\coordinate (gend) at ($(xhat)+(\gdeg:3.05)$);
\draw[->, ultra thick, cb_green] (xhat) -- (gend);
\node[font=\small\bfseries, anchor=west, cb_green]
  at ($(xhat)+(1.8,1.0)$)
  {$-\nabla_{\hat{x}} V_\theta(\hat{x})$};

\pgfmathsetmacro{\A}{2.0}
\pgfmathsetmacro{\B}{1.0}

\pgfmathsetmacro{\sOne}{exp(\A*\rOne - \B*\rOne*\rOne)}
\pgfmathsetmacro{\sTwo}{exp(\A*\rTwo - \B*\rTwo*\rTwo)}
\pgfmathsetmacro{\sThr}{exp(\A*\rThr - \B*\rThr*\rThr)}
\pgfmathsetmacro{\smax}{max(\sOne,max(\sTwo,\sThr))}

\pgfmathsetmacro{\radMin}{0.35}
\pgfmathsetmacro{\radScale}{2.25}

\newcommand{\DrawRing}[3]{%
  \foreach \ang in {#2} {%
    \pgfmathsetmacro{\score}{exp(\A*#1*cos(\ang-\gdeg) - \B*#1*#1)}%
    \pgfmathsetmacro{\p}{\score/\smax}%
    \pgfmathsetmacro{\rad}{\radMin + \radScale*sqrt(\p)}%
    \filldraw[black, fill=#3] ($(xhat)+(\ang:#1)$) circle[radius=\rad pt];
  }%
}

\DrawRing{\rOne}{0,30,60,90,120,150,180,210,240,270,300,330}{cb_blue}
\DrawRing{\rTwo}{15,45,75,105,135,165,195,225,255,285,315,345}{cb_blue}
\DrawRing{\rThr}{0,45,90,135,180,225,270,315}{cb_blue}

\node[draw, rounded corners, fill=white, inner sep=6pt,
      font=\small, anchor=north, align=left]
  at (\W/2 -0.7,\H+0.10)
{$q_\mathrm{mixing}(x\to y)\propto \exp\bigl(
- \lambda^L_{\mathcal{V}}\|\hat{y}_{\mathcal{V}}-\hat{x}_{\mathcal{V}}\|_2^2
- \lambda^L_{\mathcal{E}}\|\hat{y}_{\mathcal{E}}-\hat{x}_{\mathcal{E}}\|_2^2$\\[4pt]
$\quad\quad\quad\quad\quad\quad - \beta^L\,\nabla V_\theta(\hat{x})^\top(\hat{y}-\hat{x})\bigr)$};

\node[font=\small, anchor=west] at (5.6,1.9)
{dot area $\propto\, q_\mathrm{mixing}(x\to y)$};

\end{tikzpicture}%
}
\vspace{\dimexpr-0.8em+1pt\relax}
\captionsetup{font=footnotesize}
\caption{\textbf{Proposal Scoring.} Local edits are scored by $q_\mathrm{mixing}(x\to y)$; dot area encodes proposal probability. Gradient-aligned, shorter jumps are favored, and stay proposals are resampled.}
\label{fig:proposal-middle-softmax-alpha-half}
\end{wrapfigure}

\paragraph{Learnable Components.}
The primary learnable component of our framework is the scalar energy potential $V_\theta(x)$, implemented via a neural network. While we evaluate $V_\theta$ exclusively on discrete graphs $x \in \mathcal{X}$, we leverage the continuous embedding representation $\hat{x}$ to enable gradient-based computations. Specifically, the differentiability of $V_\theta(x):=\tilde{V}_\theta(\hat{x})$ with respect to $\hat{x}$ allows us to compute gradients that directly inform and guide local jump proposals used by the sampler (see~\Cref{proposals,proposals_lang}). Thus, the continuous embedding bridges the discrete graph domain with gradient-based optimization.
\subsection{Transport-Aligned Proposal \texorpdfstring{($\epsilon\to 0$)}{(epsilon -> 0)}}
\label{proposals}
Let $(\pi_k)_{k=0}^{K}$ denote intermediate discrete distributions along a path from an initial noise distribution to the data distribution.
Given an optimal coupling $\gamma_k$ (approximated in practice by tractable methods, e.g.,~\cite{albergo2024stochastic,tong2024improving}) and its induced transport map $T_k$, we assume a consistent embedding-based representation $y = T_k(x) \iff \hat{y} = \tilde{T_k}(\hat{x})$, with $x \sim \pi_k$ and $y = T_k(x) \sim \pi_{k+1}$ denoting consecutive graph states along the transport, and $\tilde{T_k}$ a continuous map realizing $T_k$ in the embedded space.
Expressing the JKO variational objective from \eqref{jko_em} as an optimization over transport maps (see \Cref{app:jko} for derivation) and substituting the time-step parameter $\Delta t$ with the jump step-size $\eta$ yields:
\begin{equation}
\label{eq:new_T}
    T_k(x)\in\arg\min_{y\in\mathcal{X}}\left\{\frac{1}{2\eta}c(x,y)+V_\theta(y)\right\},
\end{equation}
where $\pi_{k+1}=T^\#_{k}\pi_k$ denotes the pushforward of $\pi_k$. Assuming the minimizer $T_k(x)$ \eqref{eq:new_T} lies in the interior, the first-order optimality condition becomes:
\begin{equation}
0 = \left.\nabla_{\hat{y}}\left[\frac{1}{2\eta}\tilde{c}(\hat{x},\hat{y})+\tilde{V}_\theta(\hat{y})\right]\right|_{\hat{y}=\tilde{T_k}(\hat{x})}\implies
\label{eq:optimality}
\nabla_{\hat{y}}\tilde V_\theta\!\left(\tilde T_k(\hat{x})\right)
\;=\;
-\frac{1}{2\eta}\,
\nabla_{\hat{y}}\tilde c\!\left(\hat{x},\tilde T_k(\hat{x})\right).
\end{equation}
This condition characterizes the solution and defines the training objective (\Cref{sec:training}). During sampling, given a trained potential $\tilde V_\theta$ that satisfies the optimality objective \eqref{eq:optimality}, our goal is to approximately solve \eqref{eq:new_T} using a local proposal mechanism. Instead of performing large global moves, we iteratively descend the potential by making local edits, enabling a Taylor expansion of $\tilde V_\theta(\hat y)$ around the current state $\hat x$. Rather than controlling the step-size parameter $\eta$, we simplify the (local) greedy proposal by restricting the kernel to exactly $N$ edits per jump. Since our jumps are local, we use the cost \( c_{\mathrm{loc}}(\hat x,\hat y) \) with distinct weights \(\lambda_\mathcal{V}\) and \(\lambda_\mathcal{E}\) \eqref{eq:loc-cost}.

Linearizing around $\hat{x}$: we approximate $\tilde{V}_\theta(\hat{y}) \approx \tilde{V}_\theta(\hat{x}) + (\hat{y} - \hat{x})^\top \nabla_{\hat{x}}\tilde{V}_\theta(\hat{x})$. We deterministically select the candidate $y$ from the candidate set $C_N(x)$, defined as the set of graphs reachable from $x$ via exactly $N$ \emph{discrete} modifications (edge changes/removals or node type alterations), by minimizing the sum of the potential change and the transport cost:
\begin{align}
y^*(x) = \arg &\min_{y \in C_N(x)}\left((\hat{y} - \hat{x})^\top\nabla_{\hat{x}} V_\theta(\hat{x}) + c_{\mathrm{loc}}(x,y)\right),\quad q_{\text{greedy}}(x \to y) = \mathbb{I}_{\{y=y^*(x)\}}\label{eq:greedy-proposal}
\end{align}

In the transport regime, proposed moves are deterministically projected onto directions associated with strictly negative gradients, aiming for a decrease in energy. Edits not meeting this criterion are rejected. If no candidate yields a negative gradient, the greedy sampler stalls, switching to the Langevin-based proposal $q_\mathrm{mixing}$ in \Cref{proposals_lang}. More about proposal schedules in \Cref{init_prop}.
 For further proposal implementation details, see \Cref{app:sampling-details}.

\subsection{Discrete Mixing Proposal \texorpdfstring{($\epsilon>0$)}{(epsilon > 0)}}
\label{proposals_lang}
Discretizing dynamics \eqref{em_sde} in embedding space with temperature $\epsilon$, step $\eta$, and graphs $x,y \in \mathcal{X}$:
\begin{equation}
q_{\mathrm{mixing}}(x\to y)
\,\propto\,
\exp\left(
-\frac{1}{4\epsilon\eta}\|\hat{y}-\hat{x}+\eta\nabla_{\hat{x}} V_\theta(\hat{x})\|^2
\right).
\end{equation}

We expand the mixing transition probability explicitly in terms of $\hat{y}$, simplifying by removing terms independent of $\hat{y}$ and substituting the squared-distance term $\|\hat{y}-\hat{x}\|^2$ with the local graph distance $\tilde{c}_{\mathrm{loc}}(\hat{x}, \hat{y})$. This gives:
\begin{equation}
\log q_\mathrm{mixing}(x\to y)
= -\frac{1}{4\epsilon\eta}\tilde{c}_{\mathrm{loc}}(\hat{x}, \hat{y})
- \frac{1}{2\epsilon}\nabla_{\hat{x}} V_\theta(\hat{x})^\top(\hat{y}-\hat{x}) + \text{const.}
\label{eq:mixing-transition}
\end{equation}
The proposal is effectively local due to the distance penalty, though, in principle, it can still propose an arbitrary number of edits, enabling mode escape. We define Langevin-specific parameters $(\lambda_{\mathcal{V}}^L, \lambda_{\mathcal{E}}^L, \beta^L)$ by absorbing all constants, including $\epsilon$, $\eta$, and the scaling terms $\lambda_{\mathcal{V}}, \lambda_{\mathcal{E}}$ from $c_{\mathrm{loc}}$: 
\begin{equation}
q_\mathrm{mixing}(x\to y)
\;\propto\;\exp\!\bigl(
- \lambda_{\mathcal{V}}^L\|\hat{y}_{\mathcal{V}}-\hat{x}_{\mathcal{V}}\|_2^2
- \lambda_{\mathcal{E}}^L\|\hat{y}_{\mathcal{E}}-\hat{x}_{\mathcal{E}}\|_2^2
- \beta^L\nabla_{\hat{x}} V_\theta(\hat{x})^\top(\hat{y}-\hat{x})
\bigr).
\end{equation}

The resulting kernel can be viewed as a \emph{linearized locally balanced proposal}~\citep{zanella2020informed}; related
discrete-gradient samplers appear in~\citep{grathwohl2021oops,zhang2022langevin} for particular costs or candidate sets.
Figure~\ref{fig:proposal-middle-softmax-alpha-half} illustrates the resulting geometry-aware proposal
scoring and how gradient alignment competes with distance penalties.

In practice, we pair the \( q_\mathrm{mixing} \) proposal with a \gls{mh} acceptance step to ensure convergence to the target distribution \( \pi_\theta(x) \propto \exp(-\beta_{\mathrm{mh}} V_\theta(x)) \).
In the mixing regime, the stochastic proposal is accepted with the following \gls{mh} probability:
\begin{equation}\label{eq:mh-accept}
\begin{aligned}
\alpha(x,y)
&=\min\left(1,\frac{\pi_\theta(y)\,q_{\mathrm{mixing}}(y\to x)}{\pi_\theta(x)\,q_{\mathrm{mixing}}(x\to y)}\right).\\
\end{aligned}
\end{equation}
This choice
enforces detailed balance:
\begin{equation}\label{eq:mh-detailed-balance}
\pi_\theta(x)\,q_{\mathrm{mixing}}(x\to y)\,\alpha(x,y)
=
\pi_\theta(y)\,q_{\mathrm{mixing}}(y\to x)\,\alpha(y,x)
\end{equation}
so $\pi_\theta$ is stationary for the mixing dynamics~\citep{metropolis1953equation,hastings1970monte}.
Implementation details of the proposal kernel are provided in \Cref{app:sampling-details}.  We calibrate the parameters $\beta_{\mathrm{mh}}$, $\beta^{L}$, $\lambda_{\mathcal{V}}^{L}$, and $\lambda_{\mathcal{E}}^{L}$ by maximizing the relative likelihood of generated samples under the learned energy $V_\theta$, directly after the warmup training phase described in \Cref{sec:training}.

\subsection{Initializations and Proposal Schedules}

\paragraph{Determining the Regime.}
\label{switch}
We control the regime through an energy-based binary indicator $s\in\{0,1\}$: $s=0$ selects transport (deterministic $\epsilon\to 0$ limit), while $s=1$ selects mixing (finite $\epsilon>0$). If initialized at noise, the sampler follows deterministic greedy edits guided strictly by negative gradient directions. Transition from transport to mixing occurs when the Markov chain either reaches the target energy (defined as the mean energy of the training samples estimated during training) or becomes trapped in a local minimum (i.e., no valid gradient-improving edits exist for all samples), whichever happens first. Once triggered, the sampler enters the stochastic mixing regime.

\paragraph{Sample Initialization.}
\label{init_prop}
We use two initialization regions with aligned proposal schedules:
\begin{itemize}[leftmargin=*]
\item \textbf{Noise initialization:} sample the node count from the empirical histogram, then draw node/edge
types uniformly (uniform noise). These start far from
the data distribution, so we run transport with greedy proposals $q_{\text{greedy}}$ until the energy-based switch, then enter mixing with
$q_{\mathrm{mixing}}$. 
\item \textbf{Data initialization:} draw graphs from the training set. We skip the transport phase and start in the mixing regime with $q_{\mathrm{mixing}}$, initially using a higher model temperature (lower $\beta_{\mathrm{mh}}$) to increase the acceptance rate, and then annealing (raising $\beta_{\mathrm{mh}}$) before fixed-temperature mixing (\Cref{app:sampling-details}).

\end{itemize}
Figure~\ref{fig:moses-energy-and-metrics}
summarizes the resulting energy trajectories under noise vs.\ data initialization.
Further energy diagnostics are provided in Appendix~\ref{app:energy-diagnostics}.

\begin{figure}[t!]
\centering

\begin{minipage}[t]{0.485\textwidth}
\centering
\resizebox{\linewidth}{!}{%
\begin{tikzpicture}[scale=1.15]
  \node[font=\small\bfseries] at (3.1,3.0) {Energy Trajectories};

  \begin{scope}[yshift=0.16cm]
  \draw[->, thick] (0,-0.36) -- (6.2,-0.36);
  \draw[->, thick] (0,-0.36) -- (0,2.72);
  \node[rotate=90] at (-0.7,1.18) {$V_\theta(x)$};

  \foreach \x/\lab in {0/0,1.2/100,2.4/200,3.6/300,4.8/400,6.0/500} {
    \draw (\x,-0.36) -- (\x,-0.42);
    \node[below, font=\scriptsize] at (\x,-0.42) {\lab};
  }
  \foreach \y/\lab in {-0.36/-40,0/0,0.9/100,1.8/200,2.475/275} {
    \draw (0,\y) -- (-0.08,\y);
    \node[left, font=\scriptsize] at (-0.1,\y) {\lab};
  }

  \draw[decorate, decoration={brace, amplitude=4pt, mirror}, draw=cb_teal] (0.06,0.2) -- (2.10,0.2);
  \node[font=\scriptsize, text=cb_teal] at (1.36,-0.16) {annealed-temperature mixing};
  
  \draw[decorate, decoration={brace, amplitude=4pt, mirror}, draw=cb_blue] (2.28,0.2) -- (5.94,0.2);
  \node[font=\scriptsize, text=cb_blue] at (4.20,-0.16) {fixed-temperature mixing};
  \node at (3.1,-0.92) {Inference \glsentryshort{mcmc} steps};

    \fill[cb_orange!20] (0,1.78) rectangle (6.0,2.28);
    \draw[cb_orange, line width=0.6pt] (0,2.03) -- (6.0,2.03);
    
    \fill[cb_blue!20, opacity=0.5] (0,0.18) rectangle (6.0,0.68);
    \draw[cb_blue, line width=0.6pt] (0,0.43) -- (6.0,0.43);

  \draw[decorate, decoration={brace, amplitude=4pt}, draw=cb_red] (0.06,2.34) -- (2.22,2.34);
  \node[font=\scriptsize, text=cb_red] at (1.14,2.65) {greedy proposals};

  
  \fill[cb_red!20, opacity=0.6] plot[smooth] coordinates {
    (0.0,2.392) (0.7,1.992) (1.3,1.492) (2.1,0.742)
    (2.7,0.732) (3.3,0.732) (4.2,0.732) (5.1,0.732) (6.0,0.732)
  } -- plot[smooth] coordinates {
    (6.0,0.172) (5.1,0.172) (4.2,0.172) (3.3,0.172)
    (2.7,0.172) (2.1,0.182) (1.3,0.932) (0.7,1.432) (0.0,1.832)
  } -- cycle;
  \draw[cb_red, densely dotted, line width=1.1pt] plot[smooth] coordinates {
    (0.0,2.112) (0.7,1.712) (1.3,1.212) (2.1,0.462)
    (2.7,0.452) (3.3,0.452) (4.2,0.452) (5.1,0.452) (6.0,0.452)
  };

  \fill[cb_teal!30, opacity=0.5] plot[smooth] coordinates {
    (0.0,0.74) (0.4,0.91) (0.9,0.82) (1.4,0.74)
    (2.2,0.72) (3.1,0.71) (4.2,0.71) (5.1,0.71) (6.0,0.71)
  } -- plot[smooth] coordinates {
    (6.0,0.18) (5.1,0.18) (4.2,0.18) (3.1,0.18)
    (2.2,0.18) (1.4,0.18) (0.9,0.26) (0.4,0.35) (0.0,0.18)
  } -- cycle;
  \draw[cb_teal, densely dotted, line width=1.1pt] plot[smooth] coordinates {
    (0.0,0.46) (0.4,0.63) (0.9,0.54) (1.4,0.46)
    (2.2,0.44) (3.1,0.43) (4.2,0.43) (5.1,0.43) (6.0,0.43)
  };

\begin{scope}[shift={(3.3,1.7)}]
  \draw[fill=cb_orange!20, draw=cb_orange] (0,0) rectangle (0.35,0.18);
  \draw[cb_orange, line width=0.6pt] (0.04,0.09) -- (0.31,0.09);
  \node[anchor=west, font=\scriptsize] at (0.45,0.09) {noise band};

  \draw[fill=cb_blue!20, draw=cb_blue, fill opacity=0.5] (0,-0.28) rectangle (0.35,-0.10);
  \draw[cb_blue, line width=0.6pt] (0.04,-0.19) -- (0.31,-0.19);
  \node[anchor=west, font=\scriptsize] at (0.45,-0.19) {data band};

  \draw[fill=cb_red!20, draw=cb_red] (0,-0.56) rectangle (0.35,-0.38);
  \draw[cb_red, densely dotted, line width=1.1pt] (0.04,-0.47) -- (0.31,-0.47);
  \node[anchor=west, font=\scriptsize] at (0.45,-0.47) {samples: noise initialized};

  \draw[fill=cb_teal!30, draw=cb_teal] (0,-0.84) rectangle (0.35,-0.66);
  \draw[cb_teal, densely dotted, line width=1.1pt] (0.04,-0.75) -- (0.31,-0.75);
  \node[anchor=west, font=\scriptsize] at (0.45,-0.75) {samples: data initialized};
\end{scope}

  \end{scope}
\end{tikzpicture}
}
\end{minipage}\hfill
\begin{minipage}[t]{0.485\textwidth}
\centering
\raisebox{-0.10cm}{%
\resizebox{0.985\linewidth}{!}{%
\begin{tikzpicture}[xscale=0.99, yscale=1.06]
  \node[font=\small\bfseries] at (3.5,3.35) {Validity \& Novelty Trajectories};

  \begin{scope}[yshift=-0.35cm]
  \draw[->, thick] (0,0) -- (7.2,0) node[midway, below=14pt] {Inference \glsentryshort{mcmc} steps};
  \draw[->, thick] (0,0) -- (0,3.35);
  \node[rotate=90] at (-0.9,1.6) {Fraction$\uparrow$};

  \foreach \x/\lab in {0/0,1.4/100,2.8/200,4.2/300,5.6/400,7.0/500} {
    \draw (\x,0) -- (\x,-0.06);
    \node[below, font=\scriptsize] at (\x,-0.06) {\lab};
  }
  \foreach \y/\lab in {0/0.00,0.8/0.25,1.6/0.50,2.4/0.75,2.88/0.90,3.2/1.00} {
    \draw (0,\y) -- (-0.08,\y);
    \node[left, font=\scriptsize] at (-0.1,\y) {\lab};
  }

  \draw[cb_teal, thick] plot[smooth] coordinates {
    (0.0,3.20) (0.1,1.40) (0.2,0.55) (0.3,0.26) (0.6,0.22)
    (1.0,0.30) (1.4,0.42) (1.8,0.70) (2.2,1.05) (2.6,1.40)
    (3.0,1.70) (3.4,2.00) (3.8,2.25) (4.2,2.45) (4.6,2.60)
    (5.0,2.72) (5.6,2.86) (6.2,3.06) (6.8,3.075) (7.0,3.088)
  };

  \draw[cb_teal, thick, dashed] plot[smooth] coordinates {
    (0.0,0.00) (0.1,0.05) (0.2,0.25) (0.3,0.80) (0.4,1.60)
    (0.5,2.72) (1.0,2.45) (1.5,1.70) (2.0,2.05) (2.5,2.30)
    (3.0,2.45) (3.5,2.60) (4.0,2.70) (4.5,2.78) (5.0,2.85)
    (5.5,2.90) (6.0,2.95) (6.5,2.98) (7.0,3.00)
  };

  \draw[cb_red, thick] plot[smooth] coordinates {
    (0.0,0.00) (0.6,0.08) (1.0,0.32) (1.4,1.12) (1.8,2.24)
    (2.2,2.60) (2.6,2.72) (3.0,2.80) (4.2,2.88) (5.6,2.92) (7.0,2.944)
  };
  \draw[cb_red, thick, dashed] plot[smooth] coordinates {
    (0.0,3.20) (1.4,3.187) (2.8,3.174) (4.2,3.162) (5.6,3.149) (7.0,3.136)
  };

  \begin{scope}[shift={(3.3,0.15)}]
    \draw[fill=white, draw=cb_gray] (0,0) rectangle (3.6,1.4);
    
    \draw[cb_red, thick] (0.15,1.15) -- (0.6,1.15);
    \node[anchor=west, font=\scriptsize] at (0.7,1.15) {Validity (noise initialized)};
    \draw[cb_red, thick, dashed] (0.15,0.85) -- (0.6,0.85);
    \node[anchor=west, font=\scriptsize] at (0.7,0.85) {Novelty (noise initialized)};
    
    \draw[cb_teal, thick] (0.15,0.45) -- (0.6,0.45);
    \node[anchor=west, font=\scriptsize] at (0.7,0.45) {Validity (data initialized)};
    \draw[cb_teal, thick, dashed] (0.15,0.15) -- (0.6,0.15);
    \node[anchor=west, font=\scriptsize] at (0.7,0.15) {Novelty (data initialized)};
  \end{scope}
  \end{scope}
\end{tikzpicture}
}
}
\end{minipage}
\caption{\footnotesize \textbf{Energy and Sampling Trajectories.} Left: energy evolution for noise- vs.\ data-initialized chains, with mean $\pm$ 1 std and reference energies $\mathbb{E}_{x\sim \pi_\mathrm{data}}[V_\theta(x)]$ and $\mathbb{E}_{x\sim \pi_0}[V_\theta(x)]$. Noise-initialized chains use greedy proposals to reach the data distribution, with these transport-aligned steps serving as an effective burn-in for the subsequent mixing chain; data-initialized chains use temperature annealing (low initial $\beta_{\mathrm{mh}}$, then increasing) to recover novelty. Right: validity and novelty trajectories for data-initialized (\textcolor{cb_teal}{teal}) and noise-initialized (\textcolor{cb_red}{red}) chains. Uniqueness $\approx$100\%.}
\label{fig:moses-energy-and-metrics}
\end{figure}

\subsection{Training objectives}
\label{sec:training}

\paragraph{Noise and Data Interpolation via Minibatch coupling.}
Given minibatches \( x^0 \sim \pi_0 \) and \( x^\mathrm{data} \sim \pi_\mathrm{data} \), where $\pi_0$ is the uniform noise and $\pi_{data}$ is the data distribution respectively, we pair samples using a permutation-invariant approximation of \( c_{\mathrm{FGW}}(x^0,x^\mathrm{data}) \) described in \Cref{sec:gem}. Specifically, we use histogram matching (\Cref{app:matching}) and solve for a map \( T \) induced by the minibatch coupling $\tilde{\gamma}$ using the \texttt{POT} solver \citep{Flamary2021}.
 Source graphs $x^0$ are paired with data graphs $x^\mathrm{data}=T(x^0)$.

Crucially, we utilize $c_{\mathrm{FGW}}$ solely to select optimal pairs based on global structural similarity (e.g., node and edge counts). %
Denote their embeddings by $\hat{x}^0$ and $\hat{x}^\mathrm{data}$. Define the
displacement in the embedding space as $v=\hat{x}^\mathrm{data}-\hat{x}^0$.
We sample a \emph{discrete} interpolant $x^t:=(x^t_\mathcal{V},x^t_\mathcal{E})$, where $x_\mathcal{V}^t:= [x_i^t]_{i=1}^n$ and $x_\mathcal{E}^t := [x_{ij}^t]_{i<j}$ by independently sampling each node and edge.
\begin{equation}
\resizebox{0.9\textwidth}{!}{$
\begin{aligned}
x_i^t &\sim \operatorname{Cat}\big((1-t)\,\phi_\mathcal{V}(x_i^0) + t\,\phi_\mathcal{V}(x_i^{\mathrm{data}})\big),
&
x_{ij}^t &\sim \operatorname{Cat}\big((1-t)\,\phi_\mathcal{E}(x_{ij}^0) + t\,\phi_\mathcal{E}(x_{ij}^{\mathrm{data}})\big).
\end{aligned}
$}
\end{equation}
where $\operatorname{Cat}$ denotes categorical sampling. Let $\hat{x}^t$ be the continuous embedding of the discrete graph $x^t$ via $\phi_\mathcal{V},\phi_\mathcal{E}$ (\Cref{sec:gem}). While this embedding resides in a continuous ambient space, training is conducted solely on discrete graphs $x \in \mathcal{X}$, as sampling proposals are concerning discrete jumps. The continuous embedding thus mainly facilitates gradient computation of $\nabla_{\hat{x}_t} V_\theta(\hat{x}^t)$.

\paragraph{Objectives.}
We use the optimality condition derived in \Cref{proposals} to construct a flow-like loss $\mathcal{L}_{\text{Flow}}$. Specifically, the relation $\nabla_{\hat{y}}\tilde{V}_\theta(\hat{y})=-\frac{1}{2\eta}\nabla_{\hat{y}}\tilde{c}(\hat{x},\hat{y})$ from \eqref{eq:optimality}, evaluated at $\hat{y}=\tilde{T}_k(\hat{x})$, motivates training the negative energy gradient to match the displacement vector $v$ pointing towards the data, since for $\eta=1$ and $\lambda_\mathcal{V}=\lambda_\mathcal{E}=1$, $\left.\frac{1}{2\eta}\nabla_{\hat{y}}\tilde{c}_{\mathrm{loc}}(\hat{x}^0,\hat{y})\right|_{\hat{y}=\hat{x}^{\mathrm{data}}}=\hat{x}^{\mathrm{data}}-\hat{x}^0=v$. %
The training objective minimizes the cost ($c_{\text{loc}}$) along the interpolation path:
\begin{align}
\mathcal{L}_{\text{Flow}}(\theta)
&=
\mathbb{E}_{(x^0,x^\mathrm{data})\sim\tilde{\gamma},\,t\sim\Unif([0,1])}\left[
\|\nabla_{\hat{x}_t} \tilde{V}_\theta(\hat{x}^t)+v\|_2^2
\right],\\
\mathcal{L}_{\text{CL}}(\theta)
&=
\mathbb{E}_{x^{+}\sim \pi_\mathrm{data}}[\tilde{V}_\theta(\hat{x}^{+})]
-\mathbb{E}_{x^{-}\sim \text{sg}\,(\pi_\theta)}[\tilde{V}_\theta(\hat{x}^{-})].
\label{eq:cl}
\end{align} 

The contrastive loss $\mathcal{L}_{\text{CL}}$~\cite{hinton2002training} (see \Cref{app:energy-refinement} for the derivation) leverages the discrete proposals introduced in \Cref{proposals} and \Cref{proposals_lang} to efficiently sample from the model distribution.
The main objective is to minimize $\min_\theta\; \mathcal{L}_{\text{Flow}}(\theta)+\lambda_{\text{CL}}\,\mathcal{L}_{\text{CL}}(\theta)$. We perform a warm-up phase of $N_{\mathrm{warmup}}$ training iterations using only $\mathcal{L}_{\text{Flow}}$ before introducing $\mathcal{L}_{\text{CL}}$. This yields higher-quality negatives for contrastive learning, consistent with the continuous formulation~\citep{balcerak2025energy}. After warm-up, we temporarily freeze the network and tune the mixing-phase sampler hyperparameters $(\beta_{\mathrm{mh}}, \beta^L, \lambda_{\mathcal V}^L, \lambda_{\mathcal E}^L)$ by minimizing sample energy under fixed $V_\theta$, equivalently maximizing relative likelihood. To approximate samples from \( \pi_\theta \) required by \(\mathcal{L}_{\text{CL}}\), we run Markov chains initialized in equal proportions from uniform noise and data samples.
(See \Cref{app:exp-setup} for detailed hyperparameters.) \Cref{alg:gem-training,alg:gem-sampling} summarize training and sampling.

The contrastive formulation can be generalized to allow certain negative samples
to be suppressed more strongly than others. We explore this possibility for
invalid molecular graphs, with details and an ablation in \Cref{app:exp-setup}
and \Cref{tab:invalid-weight-ablation}.

\begin{center}
\centering
\begin{minipage}[t]{0.52\textwidth}
\refstepcounter{algorithm}\label{alg:gem-training}
\noindent\rule{\linewidth}{0.8pt}\par
\textbf{Algorithm~\thealgorithm} GEM Training\par
\noindent\rule{\linewidth}{0.4pt}\par\vspace{-0.3em}
\fontsize{9.3pt}{10.8pt}\selectfont
\begin{algorithmic}[1]
\STATE {\bfseries Input:} OT Solver, $N_{\text{CL}}, N_{\text{warmup}}, \lambda_{\text{CL}}$
\FOR{$i=0,1,2,\ldots$ until $V_\theta$ converges}
    \STATE Sample $(\xzero,\xdata)\sim\tilde{\gamma}$ (via OT Solver)
    \STATE $t\sim\mathcal{U}(0,1)$, $x^t\sim \operatorname{Cat}\big((1 - t)\hat{x}^0 + t\hat{x}^{\text{data}}\big)$
    \STATE $\mathcal{L}_{\text{Flow}}=\|\nabla\tilde{V}_\theta(\hat{x}^t)+(\hat{x}^{\text{data}}-\hat{x}^0)\|^2$
    \STATE \algorithmicif\ $i>N_{\text{warmup}}$ \algorithmicthen
    \STATE \hspace{1em}$x \sim \text{GEM Sampling}(V_\theta,N_{\text{CL}})$
    \STATE \hspace{1em}$\mathcal{L}_{\text{CL}}=\mathbb{E}[\tilde{V_\theta}(\hat{x}^{\text{data}})]-\mathbb{E}[\tilde{V_\theta}(\hat{x})]$
    \STATE \algorithmicelse\ $\mathcal{L}_{\text{CL}}=0$ \algorithmicendif
    \STATE \textbf{optimizer\_step}($\mathcal{L}_{\text{Flow}}+\lambda_{\text{CL}}\mathcal{L}_{\text{CL}}$) 
\ENDFOR
\end{algorithmic}
\noindent\rule{\linewidth}{0.4pt}
\end{minipage}\hfill
\begin{minipage}[t]{0.46\textwidth}
\refstepcounter{algorithm}\label{alg:gem-sampling}
\noindent\rule{\linewidth}{0.8pt}\par
\textbf{Algorithm~\thealgorithm} GEM Sampling\par
\noindent\rule{\linewidth}{0.4pt}
\fontsize{9.3pt}{10.8pt}\selectfont

\begin{algorithmic}[1]
\STATE {\bfseries Input:} $V_\theta$, $N_{\text{CL}}$
\STATE $V_\text{th}=\mathbb{E}_{\pidata}[\Vtheta]$
\STATE Init. $x\sim\pizero$ (or $\pidata$)
\WHILE{$\Vtheta(x)>V_\text{th}$ {\bfseries and} $x$ not stuck}
    \STATE $x\sim \qgreedy(x\to y)$
\ENDWHILE
\FOR{$k=1$ {\bfseries to} $N_{\text{CL}}$}
    \STATE $x\sim \text{MH-step}(\qmixing(x\to y))$
\ENDFOR
\STATE {\bfseries Return} $x$
\end{algorithmic}
\noindent\rule{\linewidth}{0.4pt}
\end{minipage}
\end{center}

\section{Experiments}
\label{sec:experiments}

We use two established molecular datasets, QM9~\cite{ramakrishnan2014qm9} and MOSES~\cite{polykovskiy2020moses}. QM9 comprises $\sim$150K small organic molecules containing up to nine heavy atoms (excluding hydrogens), whereas MOSES is a larger, drug-like dataset with $\sim$1.5M molecules featuring up to 27 heavy atoms. We report (i) generation on both QM9 and MOSES, (ii) conditional generation (property optimization) on MOSES, and (iii) analyses of graph-to-graph geodesics on MOSES. We additionally evaluate \gls{gem} on the non-molecular Planar and SBM datasets~\citep{martinkus2022spectre} (\Cref{app:nonmolecular}).

For unconditional generation on QM9 and MOSES, we report \gls{vun} and \gls{fcd}~\citep{preuer2018frechet}. For conditional tasks, we report \gls{cvun}, measuring the proportion of generated graphs that are valid, unique, novel, and fulfill prescribed property conditions. We evaluate two initialization regimes: noise initialization, starting from easy-to-sample noise and targeting valid molecules, and data initialization, natural for EBMs, starting from training molecules and targeting novel edits. Baseline details, experimental setup, uncertainty estimates, hyperparameter sensitivity, and wall-clock measurements are provided in \Cref{app:baselines,app:exp-setup}.

\begin{table}[t]
\centering
\footnotesize
\captionsetup{skip=3pt}
\captionsetup[subtable]{position=top,skip=3pt,font=bf}
\begin{subtable}[t]{0.48\textwidth}
\centering
\subcaption{MOSES ($\sim$1.5M molecules).}
\label{tab:uncond-moses}
\resizebox{\linewidth}{!}{%
\begin{tabular}{lcc}
\toprule
Method & \acrshort{vun} $\uparrow$ & \acrshort{fcd} $\downarrow$ \\
\midrule
Training samples (MOSES) & 0.0 & 0.25 \\
\midrule
\multicolumn{3}{l}{\textit{Noise Initialization}} \\
\midrule
GraphEBM$^*$~\citep{liu2021graphebm} & 0.081 & 9.83 \\
\gls{vfm}$^*$~\citep{eijkelboom2024vfm} & 0.814 & 2.71 \\
DeFoG$^*$~\citep{qinmadeira2024defog} & 0.822 & 1.95 \\
\textbf{\gls{gem} (Ours)} & \textbf{0.923} & \textbf{1.39} \\
\midrule
\multicolumn{3}{l}{\textit{Data Initialization}} \\
\midrule
GraphEBM$^*$~\citep{liu2021graphebm} & 0.322 & 7.31 \\
\textbf{\gls{gem} (Ours)} & \textbf{0.898} & \textbf{0.76} \\
\bottomrule
\end{tabular}
}
\end{subtable}
\hfill
\begin{subtable}[t]{0.47\textwidth}
\centering
\subcaption{QM9 ($\sim$0.15M molecules).}
\label{tab:uncond-qm9}
\resizebox{\linewidth}{!}{%
\begin{tabular}{lcc}
\toprule
Method & \glsentryshort{vun} $\uparrow$ & FCD $\downarrow$ \\
\midrule
Training samples (QM9) & 0.000 & 0.04 \\
\midrule
\multicolumn{3}{l}{\textit{Noise Initialization}} \\
\midrule
GraphEBM$^*$~\citep{liu2021graphebm} & 0.079 & 7.16 \\
\gls{vfm}~\citep{eijkelboom2024vfm} & 0.489 & \textbf{0.44} \\
DeFoG~\citep{qinmadeira2024defog} & 0.348 & 0.81 \\
\textbf{\gls{gem} (Ours)} & \textbf{0.612} & 0.92 \\
\midrule
\multicolumn{3}{l}{\textit{Data Initialization}} \\
\midrule
GraphEBM$^*$~\citep{liu2021graphebm} & 0.103 & 5.15 \\
\textbf{\gls{gem} (Ours)} & \textbf{0.502} & \textbf{0.31} \\
\bottomrule
\end{tabular}
}
\end{subtable}
\caption{\footnotesize \textbf{Unconditional Generation.} Results by initialization for (a) MOSES and (b) QM9. Both report \acrshort{vun} and \acrshort{fcd}; higher \acrshort{vun} and lower \acrshort{fcd} are better. $^*$Matched DeFoG backbone; \glspl{ebm} use a small energy head (\Cref{app:perm-invariance}). MOSES: 25k generations, 1000 steps; QM9: 10k generations, 500 steps.}
\label{tab:uncond-main}
\vspace{-0.6em}
\end{table}

\subsection{Unconditional Generation}

\begin{wrapfigure}[14]{r}{0.50\textwidth}
\vspace{-3.0em}
\centering
\resizebox{\linewidth}{!}{%
\begin{tikzpicture}[x=1.08cm,y=0.95cm]
  \draw[->, thick] (0,0) -- (6.2,0) node[midway, below=8pt] {Inference steps};
  \draw[->, thick] (0,0) -- (0,4.0);
  \node[rotate=90] at (-0.95,2.0) {\glsentryshort{vun}$\uparrow$};
  \draw[->, thick] (6.2,0) -- (6.2,4.0);
  \node[rotate=90] at (7.1,2.0) {\glsentryshort{fcd}$\downarrow$};

  \node[font=\small\bfseries] at (3.1,4.35) {Performance vs Inference Steps};

  \foreach \x/\lab in {0/0,2.067/500,4.133/1000,6.200/1500} {
    \draw (\x,0) -- (\x,-0.06);
    \node[below, font=\scriptsize] at (\x,-0.06) {\lab};
  }
  \foreach \y/\lab in {0/0.0,1.0/0.25,2.0/0.50,3.2/0.80,3.4/0.85,3.6/0.90,4.0/1.00} {
    \draw (0,\y) -- (-0.08,\y);
    \node[left, font=\scriptsize] at (-0.1,\y) {\lab};
  }
  \foreach \y/\lab in {0/0,1.0/2,2.0/4,3.0/6,4.0/8} {
    \draw (6.2,\y) -- (6.28,\y);
    \node[right, font=\scriptsize] at (6.28,\y) {\lab};
  }
  \draw[blue!70!black, thick] plot coordinates {
    (0.000,0.000) (0.413,0.100) (0.827,2.656) (0.930,3.168) (1.343,3.632)
    (1.757,3.660) (2.170,3.684) (2.583,3.700) (2.997,3.696) (3.410,3.696)
    (3.823,3.696) (4.237,3.692) (4.650,3.696) (5.063,3.692) (5.477,3.684)
    (5.890,3.688) (6.200,3.696)
  };

  \draw[orange!90!black, thick] plot coordinates {
    (0.000,0.000) (0.413,1.772) (0.827,3.476) (1.240,3.540) (1.653,3.548)
    (2.067,3.548) (2.480,3.544) (2.893,3.556) (3.307,3.552) (3.720,3.568)
    (4.133,3.556) (4.547,3.548) (4.960,3.560) (5.373,3.560) (5.787,3.564)
    (6.200,3.560)
  };

  \draw[green!60!black, thick] plot coordinates {
    (0.000,0.000) (0.413,3.220) (0.827,3.270) (1.240,3.280) (1.653,3.284)
    (2.067,3.296) (2.480,3.313) (2.893,3.278) (3.307,3.304) (3.720,3.296)
    (4.133,3.312) (4.547,3.301) (4.960,3.292) (5.373,3.308) (5.787,3.279)
    (6.200,3.298)
  };

  \begin{scope}
    \clip (0,0) rectangle (6.2,4.0);
    \draw[blue!70!black, thick, dash dot] plot coordinates {
      (0.413,6.7955) (0.827,4.1195) (0.930,4.1120) (1.343,1.6325) (1.757,1.1890)
      (2.170,0.9665) (2.583,0.8480) (2.997,0.7920) (3.410,0.7300) (3.823,0.7025)
      (4.237,0.6950) (4.650,0.6670) (5.063,0.6610) (5.477,0.6600) (5.890,0.6600)
      (6.200,0.6670)
    };
    \draw[orange!90!black, thick, dash dot] plot coordinates {
      (0.000,0.122) (0.413,0.895) (0.827,0.520) (1.240,0.412) (1.653,0.338)
      (2.067,0.329) (2.480,0.328) (2.893,0.323) (3.307,0.328) (3.720,0.319)
      (4.133,0.332) (4.547,0.334) (4.960,0.333) (5.373,0.332) (5.787,0.347)
      (6.200,0.344)
    };
    \draw[green!60!black, thick, dash dot] plot coordinates {
      (0.000,2.575) (0.413,1.494) (0.827,1.121) (1.240,1.036) (1.653,1.032)
      (2.067,0.960) (2.480,1.041) (2.893,1.018) (3.307,1.030) (3.720,1.076)
      (4.133,0.980) (4.547,1.051) (4.960,1.059) (5.373,1.004) (5.787,1.059)
      (6.200,1.018)
    };
  \end{scope}

  \begin{scope}[shift={(2.25,1.35)}]
    \draw[blue!70!black, thick] (0.15,1.65) -- (0.6,1.65);
    \node[anchor=west, font=\scriptsize] at (0.7,1.65) {\glsentryshort{gem} (noise init.) \glsentryshort{vun}};
    \draw[orange!90!black, thick] (0.15,1.38) -- (0.6,1.38);
    \node[anchor=west, font=\scriptsize] at (0.7,1.38) {\glsentryshort{gem} (data init.) \glsentryshort{vun}};
    \draw[green!60!black, thick] (0.15,1.11) -- (0.6,1.11);
    \node[anchor=west, font=\scriptsize] at (0.7,1.11) {DeFoG \glsentryshort{vun}};
    \draw[blue!70!black, thick, dash dot] (0.15,0.80) -- (0.6,0.80);
    \node[anchor=west, font=\scriptsize] at (0.7,0.80) {\glsentryshort{gem} (noise init.) \glsentryshort{fcd}};
    \draw[orange!90!black, thick, dash dot] (0.15,0.53) -- (0.6,0.53);
    \node[anchor=west, font=\scriptsize] at (0.7,0.53) {\glsentryshort{gem} (data init.) \glsentryshort{fcd}};
    \draw[green!60!black, thick, dash dot] (0.15,0.26) -- (0.6,0.26);
    \node[anchor=west, font=\scriptsize] at (0.7,0.26) {DeFoG \glsentryshort{fcd}};
  \end{scope}
\end{tikzpicture}%
}
\vspace{-1.0em}
\captionsetup{font=footnotesize}
\caption{\textbf{\acrshort{vun} and \acrshort{fcd} vs Steps.} \acrshort{vun} (higher is better) and \acrshort{fcd} (lower is better) versus inference steps on MOSES for \gls{gem} noise initialization (uniform) with greedy warmup proposal, \gls{gem} data initialization with annealed proposal, and DeFoG (marginal).}
\label{fig:moses-vun-steps}
\vspace{-1.0em}
\end{wrapfigure}

\paragraph{MOSES.}
~\Cref{tab:uncond-main} (left) reports MOSES results. Across both noise and data initialization,
\gls{gem} outperforms baselines in terms of \gls{vun} and \gls{fcd}.
\Cref{fig:moses-vun-steps} contrasts performance versus inference steps and initialization strategies, respectively. 
At larger inference budgets, \gls{gem} substantially surpasses diffusion, yielding notably higher \gls{vun} and improved distributional fidelity (lower \gls{fcd}).
Wall-clock measurements show the same qualitative trend (\Cref{tab:wallclock-runtime}). The ablation results in \Cref{tab:transport-mixing-ablation} show that the transport-aligned phase, although an approximation of the optimal transport, is crucial in practice to move chains toward data-like high-probability regions.
See \Cref{app:published-moses} for additional baselines.
 
\paragraph{QM9.} ~\Cref{tab:uncond-main} (right) shows results on QM9. Because QM9 is small, generalizing to novel samples is difficult, and \gls{fcd} can be affected by novelty variance. \gls{gem} achieves the highest \gls{vun} under both initializations and the best \gls{fcd} under data initialization. Under noise initialization, \gls{gem} has higher novelty than \gls{vfm} (0.63 vs.\ 0.49), a tradeoff reflected in higher \gls{fcd}.
Individual metrics are reported in \Cref{app:published-moses}.

\begin{table*}[t]
\centering
\small
\captionsetup{skip=3pt}
\textbf{MOSES} ($\sim$1.5M molecules)\\[2pt]
\begingroup
\setlength{\tabcolsep}{3pt}
\resizebox{\textwidth}{!}{%
\begin{tabular}{lcc cc cc cc}
\toprule
Method
& \multicolumn{2}{c}{Condition: logS $\ge$ -2.25}
& \multicolumn{2}{c}{Condition: \acrshort{qed} $\ge$ 0.9}
& \multicolumn{2}{c}{Condition: \acrshort{tpsa} $\le$ 50}
& \multicolumn{2}{c}{Condition: logP $\le$ 1.5} \\
\cmidrule(lr){2-3} \cmidrule(lr){4-5} \cmidrule(lr){6-7} \cmidrule(lr){8-9}
& \glsentryshort{cvun} $\uparrow$ & \glsentryshort{cvu} $\uparrow$
& \glsentryshort{cvun} $\uparrow$ & \glsentryshort{cvu} $\uparrow$
& \glsentryshort{cvun} $\uparrow$ & \glsentryshort{cvu} $\uparrow$
& \glsentryshort{cvun} $\uparrow$ & \glsentryshort{cvu} $\uparrow$ \\
\midrule
Training samples (MOSES) & 0.0 & 0.24 & 0.0 & 0.16 & 0.0 & 0.21 & 0.0 & 0.15 \\
\midrule
\multicolumn{9}{l}{\textit{Noise Initialization}} \\
\midrule
G-\gls{vfm}$^*$~\citep{eijkelboom2025controlledvfm} & 0.72 & 0.75 & 0.26 & 0.28 & 0.72 & 0.80 & 0.70 & 0.74 \\
DeFoG$^*$~\citep{qinmadeira2024defog} & 0.74 & 0.79 & 0.26 & 0.31 & 0.77 & \textbf{0.82} & 0.78 & 0.82 \\
\textbf{\gls{gem} (Ours)} & \textbf{0.80} & \textbf{0.81} & \textbf{0.40} & \textbf{0.40} & \textbf{0.80} & 0.81 & \textbf{0.85} & \textbf{0.85} \\
\bottomrule
\end{tabular}
}
\endgroup
\caption{\footnotesize \textbf{Property Optimization.} MOSES conditional generation under property constraints. \gls{cvun} is the fraction meeting the condition and valid, unique, novel; \gls{cvu} is the fraction meeting the condition and valid, unique. 1k inference steps, 5k samples. $^*$ Reproduced with matched backbones (\Cref{app:perm-invariance}).}
\label{tab:moses-conditional}
\vspace{-0.6em}
\end{table*}

\begin{figure*}[t]
\centering
\resizebox{\textwidth}{!}{%
\begin{tikzpicture}
  \def\boxsize{2.5}
  \def\gap{2.9}
  \def\rowgap{2.45}
  \def\xoffset{2.35}
  \def\arrowlen{0.34375} 

  \node[inner sep=0pt] at (\xoffset+0*\gap+0.5*\boxsize,0.5*\boxsize)
    {\includegraphics[width=\boxsize cm]{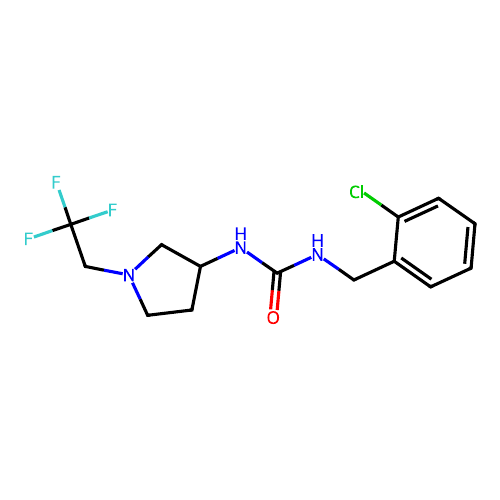}};
  \node[inner sep=0pt] at (\xoffset+1*\gap+0.5*\boxsize,0.5*\boxsize)
    {\includegraphics[width=\boxsize cm]{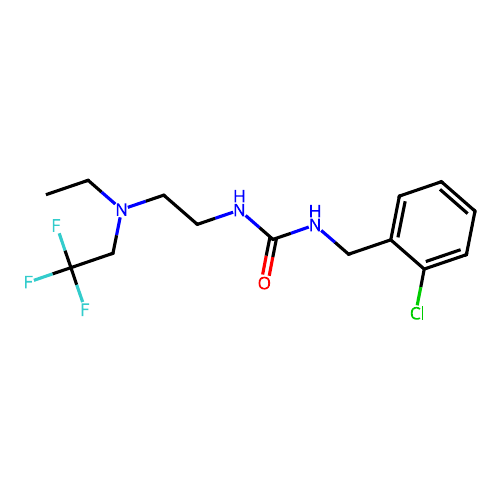}};
  \node[inner sep=0pt] at (\xoffset+2*\gap+0.5*\boxsize,0.5*\boxsize)
    {\includegraphics[width=\boxsize cm]{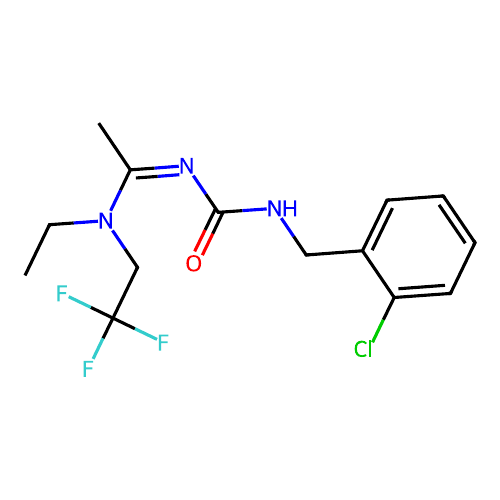}};
  \node[inner sep=0pt] at (\xoffset+3*\gap+0.5*\boxsize,0.5*\boxsize)
    {\includegraphics[width=\boxsize cm]{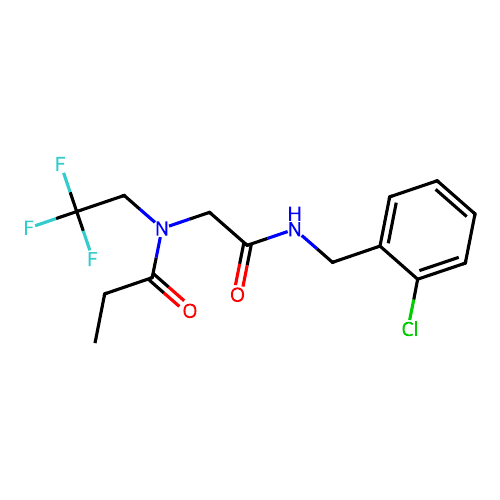}};
  \node[inner sep=0pt] at (\xoffset+4*\gap+0.5*\boxsize,0.5*\boxsize)
    {\includegraphics[width=\boxsize cm]{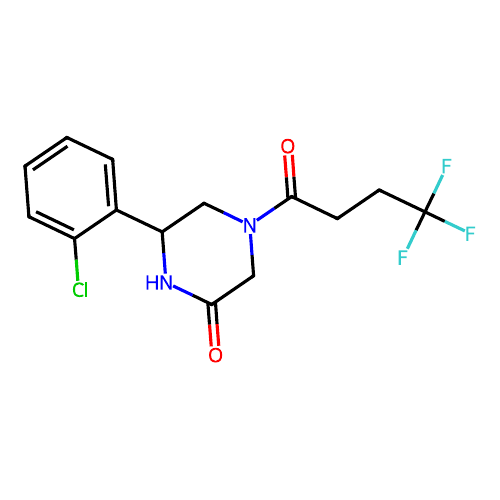}};
  \node[inner sep=0pt] at (\xoffset+5*\gap+0.5*\boxsize,0.5*\boxsize)
    {\includegraphics[width=\boxsize cm]{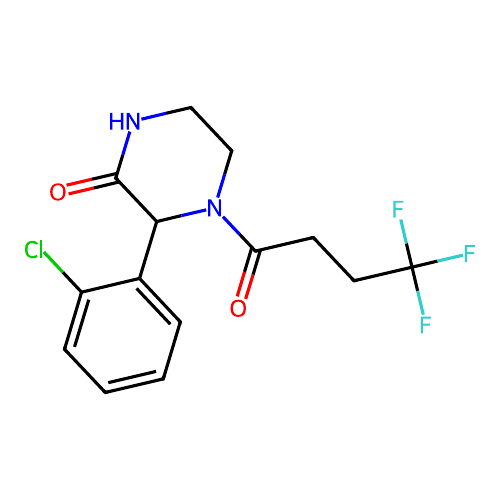}};
  \node[anchor=north west, font=\scriptsize\bfseries] at (\xoffset+0*\gap+0.05*\boxsize,0.98*\boxsize)
    {Molecule A};
  \node[anchor=north west, font=\scriptsize\bfseries] at (\xoffset+5*\gap+0.05*\boxsize,0.98*\boxsize)
    {Molecule B};
  \foreach \i in {0,1,2,3,4} {
    \draw[<->, line width=0.6pt, opacity=0.8]
      ({\xoffset+\i*\gap + 0.5*(\gap+\boxsize) - 0.5*\arrowlen}, {0.5*\boxsize})
      -- ({\xoffset+\i*\gap + 0.5*(\gap+\boxsize) + 0.5*\arrowlen}, {0.5*\boxsize});
  }

  \node[inner sep=0pt] at (\xoffset+0*\gap+0.5*\boxsize,-\rowgap+0.5*\boxsize)
    {\includegraphics[width=\boxsize cm]{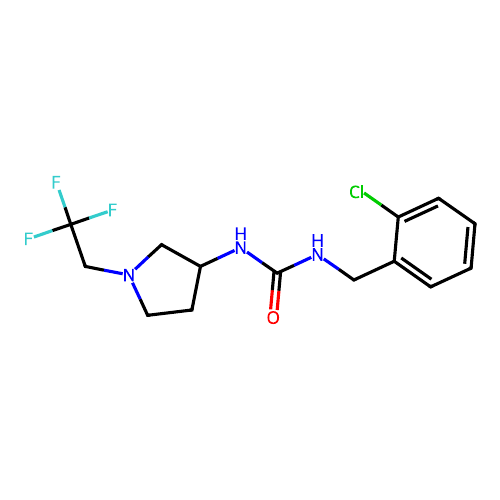}};
  \node[inner sep=0pt] at (\xoffset+1*\gap+0.5*\boxsize,-\rowgap+0.5*\boxsize)
    {\includegraphics[width=\boxsize cm]{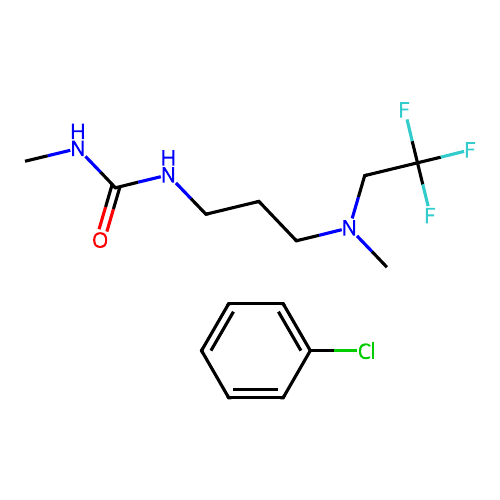}};
  \node[inner sep=0pt] at (\xoffset+2*\gap+0.5*\boxsize,-\rowgap+0.5*\boxsize)
    {\includegraphics[width=\boxsize cm]{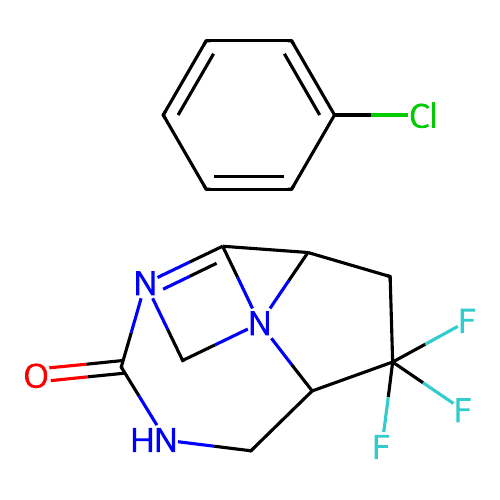}};
  \node[inner sep=0pt] at (\xoffset+3*\gap+0.5*\boxsize,-\rowgap+0.5*\boxsize)
    {\includegraphics[width=\boxsize cm]{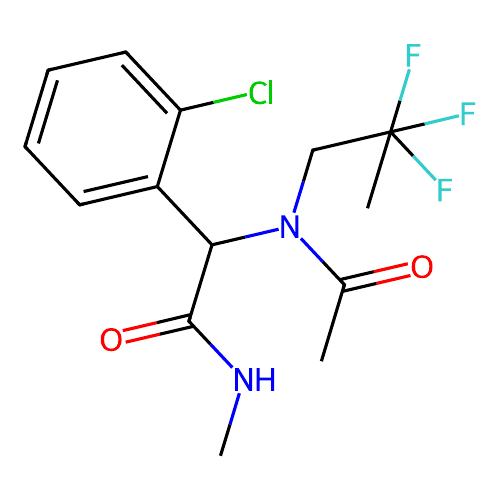}};
  \node[inner sep=0pt] at (\xoffset+4*\gap+0.5*\boxsize,-\rowgap+0.5*\boxsize)
    {\includegraphics[width=\boxsize cm]{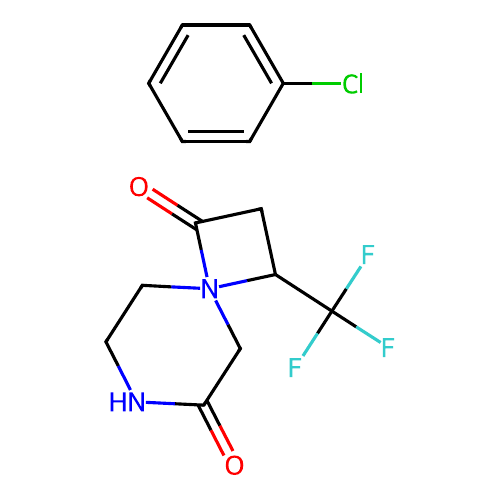}};
  \node[inner sep=0pt] at (\xoffset+5*\gap+0.5*\boxsize,-\rowgap+0.5*\boxsize)
    {\includegraphics[width=\boxsize cm]{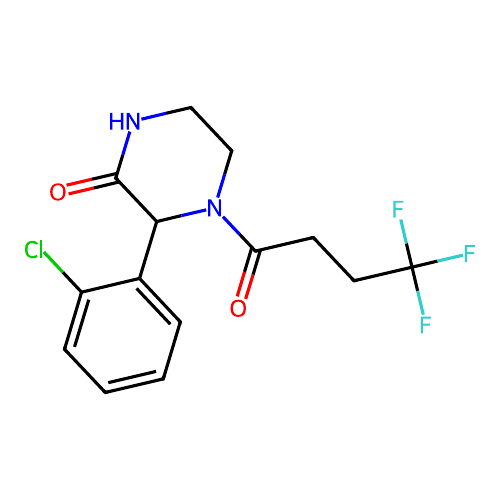}};
  \node[anchor=north west, font=\scriptsize\bfseries] at (\xoffset+0*\gap+0.05*\boxsize,-\rowgap+0.98*\boxsize)
    {Molecule A};
  \node[anchor=north west, font=\scriptsize\bfseries] at (\xoffset+5*\gap+0.05*\boxsize,-\rowgap+0.98*\boxsize)
    {Molecule B};
  \foreach \i in {0,1,2,3,4} {
    \draw[<->, line width=0.6pt, opacity=0.8]
      ({\xoffset+\i*\gap + 0.5*(\gap+\boxsize) - 0.5*\arrowlen}, {-\rowgap+0.5*\boxsize})
      -- ({\xoffset+\i*\gap + 0.5*(\gap+\boxsize) + 0.5*\arrowlen}, {-\rowgap+0.5*\boxsize});
  }

  \node[rotate=90, anchor=center, font=\scriptsize\bfseries] at (\xoffset-0.45,0.5*\boxsize) {\glsentryshort{gem} geodesic};
  \node[rotate=90, anchor=center, font=\scriptsize\bfseries] at (\xoffset-0.45,-\rowgap+0.5*\boxsize) {Cost-only geodesic};
\end{tikzpicture}}
\caption{\footnotesize \textbf{Qualitative Geodesic Paths.} Representative MOSES interpolation from Molecule A to Molecule B. Columns show successive samples along each path; \glsentryshort{gem} uses the learned energy-weighted geometry, while the baseline uses only graph-edit cost.}
\label{fig:mol-trajectories}
\vspace{-1em}
\end{figure*}

\subsection{Conditional Generation}
We optimize molecular properties (logP, logS, \gls{qed}, and \gls{tpsa}) by following~\citep{vignac2022digress,ninniri2025graph}: we train a regressor \(f_\phi\) and sample from \(V_\theta^{\mathrm{cond}}(x)=V_\theta(x)+\lambda_{\mathrm{prop}}\|f_\phi(x)-\zeta\|^2\). We run the same proposal kernel using \(-\nabla_{\hat{x}} V_\theta^{\mathrm{cond}}(\hat{x})\), with the \gls{mh} ratio modified accordingly. For fairness, all methods in \Cref{tab:moses-conditional} use noise-conditioned regressors and comparable tuning; thresholds and implementation details are in \Cref{app:exp-setup,app:regressor-noise}.
We observe that \gls{gem} maintains strong validity, uniqueness, and novelty while improving constraint satisfaction relative to baselines (\Cref{tab:moses-conditional}).

\subsection{Geodesic Analysis}
\begin{wrapfigure}[12]{r}{0.50\textwidth}
\vspace{-1.8em}
\centering
\resizebox{\linewidth}{!}{%
\begin{tikzpicture}[scale=0.87]
  \draw[->, thick] (8.5,0) -- (16.5,0) node[below, xshift=-3cm, yshift=-0.35cm] {distance (node-matching, $c_{\mathrm{FGW}}$)};
  \draw[->, thick] (8.5,0) -- (8.5,3.5);
  \node[rotate=90] at (7.8,1.75) {avg.\ validity (\%)$\uparrow$};

  \foreach \y/\label in {0/0, 0.875/25, 1.75/50, 2.625/75} {
    \draw (8.5,\y) -- (8.42,\y);
    \node[left,font=\scriptsize] at (8.4,\y) {\label};
    
    \ifdim \y pt > 0pt
        \draw[gray!40, dashed] (8.5,\y) -- (16.2,\y);
    \fi
  }

  \foreach \x in {9,12,16} {
    \draw (\x,0) -- (\x,-0.08);
    \node[below,font=\scriptsize] at (\x,-0.1) {\x};
  }

  \draw[very thick, blue] plot coordinates {(9,2.24) (12,1.435) (16,1.4)};
  \draw[very thick, gray] plot coordinates {(9,1.225) (12,0.91) (16,1.19)};

  \node[blue, anchor=west, font=\footnotesize] at (11.5,2.9) {\gls{gem} energy-weighted geodesic};
  \node[gray, anchor=west, font=\footnotesize] at (11.5,2.5) {cost-only geodesic};
\end{tikzpicture}%
}
\vspace{-1.0em}
\captionsetup{font=footnotesize}
\caption{\textbf{Validity Along Geodesics.} Average chemical validity (\%) vs.\ distance (node-matching, $c_{\mathrm{FGW}}$) for \gls{gem} energy-weighted versus cost-only geodesics. Cost-only ignores the data geometry and interpolates through low-validity regions.}
\label{fig:geodesic-validity}
\vspace{-1.0em}
\end{wrapfigure}

We study molecule-to-molecule interpolation by computing geodesics (low-energy graph-edit paths aligned with the learned data geometry \citep{bethune2025follow}). We report average validity versus the endpoint distance with distance computed via the node-matching approximation to $c_{\mathrm{FGW}}$ (\Cref{app:matching}). \Cref{fig:mol-trajectories} shows qualitative paths. As seen in \Cref{fig:geodesic-validity}, energy-weighted geodesics preserve validity more reliably than cost-only paths, indicating that the energy encodes chemically meaningful structure. See~\Cref{app:geodesic-details} for experimental details.

\section{Discussion and Conclusion}
We introduced \emph{Graph Energy Matching} (\gls{gem}), a discrete generative framework that extends continuous energy matching principles to discrete graph domains motivated by the transport-map formulation of the \gls{jko} scheme. \gls{gem} integrates deterministic, transport-aligned graph edits with effective mixing under a unified scalar energy model. The transport stage serves primarily as an effective burn-in mechanism that guides sampling toward high-probability regions. Conceptually, this allows (i) scalable and high-quality molecular graph generation from both noise- and data-initialized states, and (ii) flexible conditional generation at inference-time through an explicitly learned relative-likelihood structure. Empirically, \gls{gem} achieves state-of-the-art performance among graph \glspl{ebm}, matching or surpassing discrete diffusion baselines on most reported MOSES and QM9 metrics. While our empirical evaluation focuses primarily on molecular graphs, \gls{gem} also remains highly competitive on the non-molecular Planar and SBM datasets. Further, it effectively addresses property optimization under threshold constraints. Moreover, \gls{gem}'s representation naturally enables diagnostics such as graph geodesics.

\paragraph{Broader Context and Future Directions.}
Our current formulation operates in a one-hot embedding space, where distinct categories at each node or edge are equally separated. Although the learned energy distinguishes between edits, the embedding geometry itself does not encode similarities between categories. Learning more compact representations that capture these similarities could provide a more informative geometry for transport and proposal construction. More broadly, graphs are a particular class of structured discrete objects. Extending \gls{gem}'s transport--mixing formulation to other discrete object distributions would require suitable representations, edit operations, and transport costs. Exploring how the same scalar energy can support transport and mixing across these domains is an important direction for future work.

\paragraph{Limitations.}
\gls{gem} learns a scalar energy field capturing relative likelihood. Consequently, like other \glspl{ebm}, generation velocities must be derived from gradients, incurring a $\sim$2--3$\times$ theoretical overhead compared to purely forward parameterizations.

\FloatBarrier
\section*{Acknowledgments}
We thank Antonio Terpin for helpful discussions. This work has been made possible in part by a gift from the Chan Zuckerberg Initiative Foundation to establish the Kempner Institute for the Study of Natural and Artificial Intelligence. Support came from the Helmut Horten Foundation and the European Cooperation in Science and Technology (COST). We declare no competing interests.

{\small
\bibliographystyle{plainnat}
\bibliography{main}
}

\appendix
\section{Permutation-invariant energy parameterization}\label{app:perm-invariance}
We implement the potential with a graph transformer backbone directly from DeFoG~\cite{qinmadeira2024defog}. Each layer of this backbone is permutation equivariant, meaning that for any permutation (relabeling) $\sigma$, the layer satisfies:
\[
h^{(l)}(\sigma \cdot x_{\mathcal{V}}^{(l)}, \sigma \cdot x_{\mathcal{E}}^{(l)}, f^{(l)}) = (\sigma \cdot x_{\mathcal{V}}^{(l+1)}, \sigma \cdot x_{\mathcal{E}}^{(l+1)}, f^{(l+1)}),
\]
where $x_{\mathcal{V}}^{(l)}$ and $x_{\mathcal{E}}^{(l)}$ denote node and edge features at layer $l$, respectively, and $f^{(l)}$ denotes the global graph-level feature.

Following~\citep{vignac2022digress,qinmadeira2024defog}, we pad and mask all smaller graphs to the maximum number of nodes observed in the training dataset, denoted as $n$. This results in fixed-size inputs to the model, denoted $x_\mathcal{V}$ and $x_\mathcal{E}$ for node and edge features, respectively. We denote node and edge features at layer $l$ as $x^{(l)}_\mathcal{V}$ and $x^{(l)}_\mathcal{E}$, respectively.
The global feature $f^{(l)}$ collects graph-level features at layer $l$ and is updated through a permutation-invariant operation.
Figure~\ref{fig:energy-arch} summarizes the backbone and invariant readout.

\begin{figure}[h]
\centering
\resizebox{\columnwidth}{!}{%
\begin{tikzpicture}[
  font=\scriptsize,
  node distance=1.0cm,
  module/.style={
    rectangle,
    rounded corners=5pt,
    minimum width=3.1cm,
    minimum height=0.95cm,
    align=center,
    draw=black!55,
    fill=black!4,
    thick
  },
  arrow/.style={-Latex, line width=1.1pt, draw=black!60}
]
  \node[module] (graph) {Input Graph \\\\ $x:=(x^{(l)}_\mathcal{V},x^{(l)}_\mathcal{E})$};
  \node[module, right=of graph] (gt) {Graph Transformer\\(Permutation Equivariant)};
  \node[module, right=of gt] (pool) {Invariant Readout\\\\$\sum_i x^{(L)}_i,\;\sum_{i\ne j} x^{(L)}_{ij},f^{(L)}$ };
  \node[module, right=of pool] (energy) {$V_\theta(x)\in\mathbb{R}$};

  \draw[arrow] (graph) -- (gt);
  \draw[arrow] (gt) -- (pool);
  \draw[arrow] (pool) -- (energy);
\end{tikzpicture}}
\caption{\textbf{Energy Backbone.} Permutation-equivariant backbone with invariant pooling to produce a scalar energy.}
\label{fig:energy-arch}
\end{figure}

The scalar energy is obtained by an invariant readout that first pools the
equivariant features and then applies a \gls{mlp}:
\begin{equation}
\begin{aligned}
\bar{x} &= \sum_{i\in\mathcal{V}} x^{(L)}_i,\qquad
\bar{e} = \sum_{i< j} x^{(L)}_{ij},\\
V_\theta(x) &= \mathrm{MLP}\big([\bar{x},\bar{e},f^{(L)}]\big).
\end{aligned}
\end{equation}
Since the sum operation is permutation invariant, the concatenated vector
$(\bar{x},\bar{e},f^{(L)})$ is invariant. Therefore, an \gls{mlp} applied to it is
also invariant.

\paragraph{Energy head and parameter count.}
To enhance expressivity, we incorporate Relative Random Walk Probabilities (RRWP)~\citep{ma2023graph}
as node and edge features, matching DeFoG for a fair comparison. The energy MLP head is a 2-layer MLP
that maps the pooled hidden graph representation (concatenated node/edge/$f$ embeddings) to a single
scalar energy. With dimensions $448\to256\to1$ and SiLU activation, it has 115,201 parameters. The
full model has 16,382,158 trainable parameters ($\approx 16.38$M).

\section{JKO reduction to a transport map}
\label{app:jko}

We specialize the continuous JKO functional \eqref{jko_em} to discrete distributions in the
transport regime ($\epsilon=0$) and replace the time step $\Delta t$ with the jump size $\eta$.
Given $\pi_k$ and a candidate $\pi$, the variational objective reads
\begin{align}
    \pi_{k+1} = \arg\min_{\pi}
    \frac{1}{2\eta}\inf_{\gamma\in\Gamma(\pi_k,\pi)}&\mathbb{E}_{(x,y)\sim\gamma}[c(x,y)]
    \nonumber\\
    + &\mathbb{E}_{y\sim\pi}[V_\theta(y)].
\end{align}
Assuming the optimal coupling admits a deterministic transport map $T$ (i.e., a Monge-type
solution) so that $\pi = T_{\#}\pi_k$, the objective reduces to an optimization over maps:
\begin{align}
    T \in \arg&\min_{T^\prime:\pi=T^\prime_{\#}\pi_k}
    \frac{1}{2\eta}\mathbb{E}_{x\sim\pi_k}[\,c(x,T^\prime(x))\,] \nonumber\\
    & + \mathbb{E}_{x\sim\pi_k}[\,V_\theta(T^\prime(x))\,].\\[5pt]
    T \in \arg&\min_{T^\prime}\mathbb{E}_{x\sim\pi_k}\left[\frac{1}{2\eta}c(x,T^\prime(x))+V_\theta(T^\prime(x))\right].
\end{align}
Since the expectation is over $x\sim\pi_k$, the minimization decouples pointwise, yielding
\begin{equation}
    T(x)\in\arg\min_{y}\left\{\frac{1}{2\eta}c(x,y)+V_\theta(y)\right\},\quad\text{for each }x\sim\pi_k.
\end{equation}
This yields the update used in
\Cref{proposals}.

\paragraph{Scope of the transport interpretation.}
This derivation is used as modeling motivation for the training objective and transport-aligned proposals. It does not imply exact optimal transport for the complete implemented sampler, which uses discrete graph edits, approximate minibatch matching, and local linearization. We therefore do not claim end-to-end formal optimality for the practical sampler.

\section{Permutation-invariant minibatch matching}
\label{app:matching}

Exact computation of the hard-permutation \gls{fgw} cost $c_{\mathrm{FGW}}$ requires a combinatorial search over
permutations. We therefore use fast, permutation-invariant hard-assignment approximations
tailored to the source distribution and the downstream use (pairing vs.\ local edits).

\paragraph{Histogram matching for a uniform source.}
When the source $\pi_0$ is a uniform distribution over node and edge classes, we use a cheap signature
for each graph. Let $h_\mathcal{V}$ be the normalized node-type histogram, $h_\mathcal{E}$ the normalized edge-type
histogram (over $i<j$), and $h_{\mathcal{V}\mathcal{E}}$ the histogram over unordered node-type pairs with edge types. We
form a weighted signature $h(x)=[\alpha_1 h_\mathcal{V},\;\alpha_2 h_\mathcal{E},\;\alpha_3 h_{\mathcal{V}\mathcal{E}}]$ and solve a linear assignment
between graphs of equal size using $\|h(x)-h(y)\|_1$. This produces permutation-invariant couplings and
is sufficient when $\pi_0$ is far from $\pi_{\mathrm{data}}$, so we do not require fine-grained alignment. $\alpha_{*}$ are chosen to balance each contribution equally.

\paragraph{Node matching for molecule-like sources.}
If the source distribution already produces valid molecules, we use a node-matching permutation alignment
based on node labels, followed by computing the
aligned local cost. This reflects chemical atom mapping practice: once atoms are aligned, bond
correspondences are largely determined~\citep{bell2019dockrmsd}.

\section{Sampling proposal implementation details}
\label{app:sampling-details}

\paragraph{Gradient features.}
We use gradient-informed proposals on the discrete graph space with target density
\(\pi_\theta(x)\propto \exp\left(-\beta_{\mathrm{mh}}V_\theta(x)\right)\). To compute the gradients of the potential, we first embed \( x \) into a continuous, real-valued one-hot representation \( \hat{x} \), and then evaluate the gradient with respect to this embedding as \( g := \nabla_{\hat{x}} V_{\theta}(\hat{x}) \).
We reshape node and edge gradients as
\begin{align}
g^\mathcal{V} & := \operatorname{reshape}_{n,l_{\mathrm{node}}}\big(\nabla_{\hat{x}_\mathcal{V}} V_\theta(\hat{x})\big),\\
g^\mathcal{E} & := \operatorname{reshape}_{\binom{n}{2},l_{\mathrm{edge}}}\big(\nabla_{\hat{x}_\mathcal{E}} V_\theta(\hat{x})\big),
\end{align}
where $\operatorname{reshape}_{m,n}(\cdot)$ is a row-major reshape into $\mathbb{R}^{m\times n}$.

Since we are working with undirected graphs, we use symmetrized edge gradients from the graph transformer.

\paragraph{Greedy transport implementation.}
One forward pass and one backward pass compute the energy gradient used to score
all categorical node and edge replacements in \eqref{eq:greedy-proposal}.
We retain the best replacement at each site, rank these candidates jointly,
and apply the top $N$ edits to construct one proposed graph without enumerating $C_N(x)$.
We use $N=1$ for the main QM9 and MOSES experiments.
The parameter $N$ controls only transport and does not fix the number of edits during mixing.
See Appendix~\ref{app:robustness-runtime} for sensitivity to $N$ and computational costs.

\paragraph{Factorized proposal (multi-site).}
We sample all nodes and undirected edges independently from categorical logits
\begin{align}
\ell^\mathcal{V}_{i,c_i} & = \beta\big(g^\mathcal{V}_{i,x_i}-g^\mathcal{V}_{i,c_i}\big)-\lambda_\mathcal{V}\,\mathbb{I}_{\{c_i \neq x_i\}};~\forall~c_i \in \mathcal{X}_\mathrm{node} \\
\ell^\mathcal{E}_{{ij},c_{ij}} &= \beta\big(g^\mathcal{E}_{ij,x_{ij}}-g^\mathcal{E}_{ij,c_{ij}}\big) \nonumber\\
&-\lambda_\mathcal{E}\,\mathbb{I}_{\{c_{ij}\ne x_{ij}\}};~\forall~c_{ij} \in \mathcal{X}_\mathrm{edge}
\end{align}
so the full proposal factorizes as
\begin{equation}
\begin{aligned}
q_{\beta}(x\to y)
&=
\prod_{i\in\mathcal{V}}\mathrm{Cat}\left(y_i;\mathrm{softmax}(\ell_i^\mathcal{V})\right)\\
&\quad\times\prod_{i<j}\mathrm{Cat}\left(y_{ij};\mathrm{softmax}(\ell_{ij}^\mathcal{E})\right).
\end{aligned}
\end{equation}
$\lambda_{\mathcal{V}}$ and $\lambda_{\mathcal{E}}$ penalize modifications to nodes and edges, respectively.

\paragraph{\gls{mh} refinement (single beta).}
For the fixed-temperature proposal we set \(\beta^L=\beta_{\mathrm{mh}}=\beta\) and accept
\begin{equation}
\begin{aligned}
\alpha(x,y)
&=\min\{1,\exp(\Delta(x,y))\},\\
\Delta(x,y)
&=-\beta_{\mathrm{mh}}\big(V_\theta(y)-V_\theta(x)\big)
+\log\frac{q_{\beta^L}(y\to x)}{q_{\beta^L}(x\to y)}.
\end{aligned}
\end{equation}

\paragraph{Guarantee scope.}
The formal Markov-chain guarantee applies to the fixed-temperature \gls{mh} mixing stage: with the acceptance ratio above, the chain satisfies detailed balance with target \(\pi_\theta\), so \(\pi_\theta\) is stationary. The preceding transport stage and optional temperature annealing are practical burn-in mechanisms that accelerate entry into high-probability regions before stationary mixing.

\paragraph{Two-betas simulated annealing.}
We decouple the proposal temperature and target temperature by using
\(\beta^L\) in the logits above, while \(\beta_{\mathrm{mh}}\) controls the \gls{mh} ratio.
Optionally, \(\beta_{\mathrm{mh}}\) is annealed across $S_{\mathrm{anneal}}$ mixing steps:
\begin{equation}
\beta_{\mathrm{mh}}(s)=
\begin{cases}
\beta_{\mathrm{mh}}^{\mathrm{init}}+\Delta\beta\,\dfrac{s}{S_{\mathrm{anneal}}-1}, & 0\le s<S_{\mathrm{anneal}},\\[6pt]
\beta_{\mathrm{mh}}^{\mathrm{final}}, & s\ge S_{\mathrm{anneal}}.
\end{cases}
\end{equation}
where \(\Delta\beta=\beta_{\mathrm{mh}}^{\mathrm{final}}-\beta_{\mathrm{mh}}^{\mathrm{init}}\).

\paragraph{Do-nothing resampling and proposal diffusion.}
If the independently sampled proposal equals the current state (all sites \textit{stay}),
we resample after softening the logits to increase randomness:
\(\beta^L\leftarrow\rho\,\beta^L\),
\(\lambda_\mathcal{V}\leftarrow\rho\,\lambda_\mathcal{V}\),
\(\lambda_\mathcal{E}\leftarrow\rho\,\lambda_\mathcal{E}\) with \(0<\rho<1\).
We repeat this a bounded number of times; if no change occurs, we keep the current
state and accumulate the \textit{stay} probability in \(\log q\). The reverse
transition uses the same resampling rule, preserving the \gls{mh} ratio.

\section{Conceptualizing baselines}
\label{app:baselines}

We summarize the baseline methods used in our experiments with an emphasis on the generation/sampling mechanism for each method. 

\paragraph{GraphEBM} ~\citep{liu2021graphebm} models molecular graphs using an energy-based model that assigns a scalar energy $E_\theta(x)$ to each graph $x$ via a graph neural network.
Unlike transport- or flow-based approaches, GraphEBM does not learn an explicit transformation from a source distribution to the data distribution. Instead, generation depends exclusively on Langevin dynamics sampling within a continuous embedding space.

\paragraph{DeFoG}~\citep{qinmadeira2024defog} adapts discrete flow matching to graphs by pairing an explicit noising path (via linear interpolation toward a simple base distribution) with a learned \gls{ctmc} denoiser: a network predicts the clean-graph posterior from an intermediate noisy graph, which in turn defines the time-dependent rate matrices used for generation.
Sampling starts from the base distribution and simulates the resulting \gls{ctmc} toward the data distribution, with the sampling step schedule largely selectable at inference time.

\paragraph{\gls{vfm}}~\citep{eijkelboom2024vfm} casts flow matching as variational inference over trajectory endpoints (the “posterior probability path”), yielding a KL-based objective that for categorical graph variables reduces to a per-component cross-entropy.
In contrast to DeFoG’s stochastic \gls{ctmc} jump dynamics, \gls{vfm} targets a deterministic continuous flow (vector field) on the probability simplex and generates by integrating this flow from the base distribution, then discretizing/sampling from the final categorical probabilities.

\paragraph{G-\gls{vfm}}~\citep{eijkelboom2025controlledvfm} extends \gls{vfm} with equivariant conditioning mechanisms to
support controllable generation.

\section{Experimental setup and hyperparameters}
\label{app:exp-setup}
\paragraph{MOSES (shared across experiments).}
Training is conducted in two stages. We first train with $\lambda_{\mathrm{CL}}=0.0$ for $330\,\text{k}$ iterations ($\text{lr}=10^{-4}$, batch size $=128$) to pretrain the potential field, enabling efficient burn-in of samples initialized from noise. We then follow with
$\lambda_{\mathrm{CL}}=0.1$ for 1000 iterations using chain length $N_{\mathrm{CL}}$ = 500 at
$\text{lr}=1\times 10^{-5}$, batch size 128. Greedy transport proposals followed by mixing with $\beta^L=9.55$, $\lambda_{\mathcal{V}}^L=0.23$, $\lambda_{\mathcal{E}}^L=1.88$,
and $\beta_{\mathrm{mh}}=\beta^L$. For the DeFoG baselines, we directly use the authors' provided checkpoint~\citep{qinmadeira2024defog}. During training, we consider two distinct setups for initializing the contrastive loss (CL) samples: either from noise or directly from data. We evenly split these initializations (50\% noise / 50\% data), except in the unconditional generation scenario (``Data Initialization'' category), where we exclusively (100\%) initialize CL samples from data.

\paragraph{Validity-aware contrastive weighting.}
We introduce validity-aware contrastive weighting controlled by a single
hyperparameter, $\lambda_{\mathrm{invalid}}$. For each training graph, we run one
chain, initialized from that graph or from noise with the same number of nodes.
The training graph and the chain's final sample form the contrastive pair.
The sums below group these minibatch pairs $(x^+,x^-)$ by whether the generated negative passes
RDKit~\citep{rdkit} sanitization and is connected (valid), or fails either check
(invalid):
\[
\mathcal{L}_{\text{CL}}=
\frac{
\sum_{\mathrm{valid}}\bigl[V_\theta(x^+)-V_\theta(x^-)\bigr]
+\lambda_{\mathrm{invalid}}\sum_{\mathrm{invalid}}\bigl[V_\theta(x^+)-V_\theta(x^-)\bigr]
}{\sum_{\mathrm{valid}}1+\lambda_{\mathrm{invalid}}\sum_{\mathrm{invalid}}1}.
\]
For $\lambda_{\mathrm{invalid}}>1$, invalid negatives receive greater weight than
valid ones, encouraging higher energy relative to their paired data graphs.
This validity-aware component of \gls{gem} uses the flexibility of energy-based
models to selectively suppress undesired samples during training.
We use $\lambda_{\mathrm{invalid}}=10$ for noise-initialized MOSES generation
(\Cref{tab:uncond-moses}) and $\lambda_{\mathrm{invalid}}=1$ for QM9.
Around 99\% of generated QM9 samples are already valid, leaving limited room for
validity improvements through additional weighting.
\Cref{tab:invalid-weight-ablation} reports the MOSES weighting ablation.

\subsection{Unconditional generation}
We report 25k MOSES samples and 10k QM9 samples. For noise initialization we use greedy transport
followed by the mixing proposal with the parameters from the training setup. For data initialization we
use an annealed schedule with $\beta^{L}=8.12$, $\beta_{\mathrm{mh}}^{\mathrm{init}}=0.18$,
$\beta_{\mathrm{mh}}^{\mathrm{final}}=13.56$, $S_{\mathrm{anneal}}=200$, $\lambda_{\mathcal{V}}^L=0.07$,
and $\lambda_{\mathcal{E}}^L=2.23$.

\subsection{Conditional generation and property optimization}
\paragraph{Noisy property regressors.}
We train time-conditioned regressors on MOSES using shared hyper-parameters for both marginal- and uniform-noise variants. Training uses batch size $1024$ for $60$ epochs with Adam ($\text{lr}=10^{-4}$, weight decay=$10^{-6}$). The regressor is a GraphTransformer with $4$ layers, $8$ attention heads, hidden dimensions $(d_x,d_e,d_y)=(256,128,128)$, feedforward dimensions $(256,128,192)$, and MLP dimensions $(X,E,y)=(256,128,256)$, SiLU activations, and a scalar property head. We apply polydec time distortion. Additional noise-sensitivity details are provided in \Cref{app:regressor-noise}.

\paragraph{Conditional generation (GEM \& DeFoG).}
For a desired property value $\zeta$, we define
\(V_\theta^{\mathrm{cond}}(x)=V_\theta(x)+\lambda_{\mathrm{prop}}\|f_\phi(x)-\zeta\|^2\)
and run the same proposal kernel using the gradient of \(V_\theta^{\mathrm{cond}}\), with the \gls{mh}
acceptance ratio modified accordingly. For all methods in \Cref{tab:moses-conditional}, we use
noise-conditioned regressors and comparable hyperparameter tuning. We feed regressor time using an
energy-based proxy that increases linearly until the chain reaches the data-distribution energy level,
then clamp it to $t=1$ (\Cref{app:regressor-noise}). The target constraints
(logS $\ge$ -2.25, logP $\le$ 1.5, \gls{qed} $\ge$ 0.9, \gls{tpsa} $\le$ 50) are selected so that
unconditional MOSES samples satisfy each constraint approximately 20\% of the time.

GEM mixing proposal parameters:
($\beta^L=9.55$, $\lambda_{\mathcal{V}}^L=0.23$, $\lambda_{\mathcal{E}}^L=1.88$) with uniform noise initialization/regressors.
DeFoG uses marginal-noise initialization/ regressors. For each task we sweep $\lambda_{\text{prop}}$
and report the best CVUN.

The reported results are computed directly using RDKit library \cite{rdkit}—not from the regressors, which serve solely for guidance purposes.

\begin{table}[t]
\centering
\footnotesize
\setlength{\tabcolsep}{4pt}
\begin{tabular}{lcc}
\toprule
Task & GEM $\lambda_{\text{prop}}$ & DeFoG $\lambda_{\text{prop}}$ \\
\midrule
logP $\le 1.5$ & 10 & 140 \\
logS $\ge -2.25$ & 16 & 130 \\
QED $\ge 0.9$ & 193 & 360 \\
TPSA $\le 50$ & 0.03 & 0.5 \\
\bottomrule
\end{tabular}
\caption{\footnotesize Hyper-parameters for noisy-regressor conditioning on MOSES.
GEM uses uniform-noise regressors; DeFoG uses marginal-noise regressors. Conditional results based on 5k samples.}
\label{tab:cond-hparams}
\end{table}

\subsection{Robustness, ablations, sensitivity, and runtime}
\label{app:robustness-runtime}
\paragraph{Multi-seed variability.}
We conducted additional multi-seed experiments with $\lambda_{\mathrm{invalid}}=1$, distinguishing training-seed variability from sampling-seed variability. On MOSES, training across five seeds yielded \gls{vun} $=0.858\pm0.013$ and \gls{fcd} $=1.51\pm0.03$. Sampling from a fixed checkpoint across five random seeds yielded \gls{vun} $=0.855\pm0.011$ and \gls{fcd} $=1.51\pm0.03$.
\begin{table}[h]
\centering
\footnotesize
\setlength{\tabcolsep}{5pt}
\begin{tabular}{lcc}
\toprule
Source of variability & \glsentryshort{vun} $\uparrow$ & \glsentryshort{fcd} $\downarrow$ \\
\midrule
Training seeds & $0.858\pm0.013$ & $1.51\pm0.03$ \\
Sampling seeds, fixed checkpoint & $0.855\pm0.011$ & $1.51\pm0.03$ \\
\bottomrule
\end{tabular}
\caption{\footnotesize MOSES unconditional variability across five seeds.}
\label{tab:seed-variability}
\end{table}

For conditional settings, \gls{cvun} (logS/\gls{qed}/\gls{tpsa}/logP) showed training-seed variability of $0.80\pm0.02$ / $0.40\pm0.02$ / $0.80\pm0.02$ / $0.85\pm0.02$. With a fixed checkpoint, sampling-seed standard deviations were $0.01$ / $0.02$ / $0.02$ / $0.02$, respectively. These results indicate low variance across both training and sampling randomness.

\paragraph{Invalid-negative weighting ablation.}
\Cref{tab:invalid-weight-ablation} compares noise-initialized MOSES generation at three
values of $\lambda_{\mathrm{invalid}}$. Increasing it from $1$ to $10$ improves
both \gls{vun} and \gls{fcd}. Increasing it further to $100$ has little effect.
\begin{table}[H]
\centering
\footnotesize
\begin{tabular}{ccc}
\toprule
$\lambda_{\mathrm{invalid}}$ & \acrshort{vun} $\uparrow$ & \acrshort{fcd} $\downarrow$ \\
\midrule
1 & 0.856 & 1.51 \\
10 & 0.923 & 1.39 \\
100 & 0.921 & 1.40 \\
\bottomrule
\end{tabular}
\caption{\footnotesize \textbf{Invalid-negative weighting.}
Noise-initialized \gls{gem} generation on MOSES.}
\label{tab:invalid-weight-ablation}
\end{table}

\paragraph{Transport/mixing ablation.}
We isolate the transport phase, mixing phase, and their switching combination on MOSES with $\lambda_{\mathrm{invalid}}=1$ under noise initialization with 1000 inference steps. Using the same checkpoint, full \gls{gem} reaches \gls{vun} $=0.856$, while transport-only reaches $0.712$ and mixing-only reaches $0.485$. Retraining with only mixing further degrades results because the lack of burn-in destabilizes contrastive learning, leading to early collapse after roughly 60 iterations and \gls{vun} $=0.21$.
\begin{table}[h]
\centering
\footnotesize
\setlength{\tabcolsep}{6pt}
\begin{tabular}{lc}
\toprule
Variant & \glsentryshort{vun} $\uparrow$ \\
\midrule
Full \glsentryshort{gem} (transport + mixing) & \textbf{0.856} \\
\hline
Transport-only, same checkpoint & 0.712 \\
Mixing-only, same checkpoint & 0.485 \\
Mixing-only, retrained & 0.21 \\
\bottomrule
\end{tabular}
\caption{\footnotesize Transport/mixing ablation on MOSES under noise initialization.}
\label{tab:transport-mixing-ablation}
\end{table}

\paragraph{Sampling hyperparameter sensitivity.}
Sampling-stage hyperparameters $(\beta_{\mathrm{mh}}, \beta^L, \lambda_{\mathcal{V}}^L, \lambda_{\mathcal{E}}^L)$ were tuned to minimize sample energy under fixed $V_\theta$. Evaluations on MOSES with $\lambda_{\mathrm{invalid}}=1$ over 100 independent tuning runs, each with 256 samples, yielded stable \gls{vun} results: $0.856$, $0.861$, and $0.848$ for the three best configurations. This indicates limited sensitivity to the final sampling hyperparameter choice.

\paragraph{Wall-clock runtime.}
We additionally report MOSES wall-clock measurements on an NVIDIA H100 with batch size 128, using $\lambda_{\mathrm{invalid}}=1$ for \gls{gem}. Consistent with \Cref{fig:moses-vun-steps}, DeFoG improves rapidly at early times, while \gls{gem} reaches stronger final \gls{vun} and \gls{fcd} at higher computational budgets.
\begin{table}[h]
\centering
\footnotesize
\setlength{\tabcolsep}{3pt}
\resizebox{0.85\linewidth}{!}{%
\begin{tabular}{lrrrrrrrr}
\toprule
Time (s) & 0 & 1 & 2 & 3 & 6 & 9 & 12 & 15 \\
\midrule
\glsentryshort{vun} (DeFoG) & 0.000 & 0.788 & 0.823 & 0.820 & 0.821 & 0.821 & 0.823 & 0.825 \\
\glsentryshort{vun} (Data-init) & 0.000 & 0.234 & 0.467 & 0.692 & 0.885 & 0.887 & 0.887 & 0.891 \\
\glsentryshort{vun} (Noise-init) & 0.000 & 0.012 & 0.056 & 0.370 & 0.836 & 0.855 & 0.861 & 0.860 \\
\glsentryshort{fcd} (DeFoG) & 5.150 & 2.495 & 1.956 & 2.039 & 2.097 & 2.016 & 2.021 & 2.036 \\
\glsentryshort{fcd} (Data-init) & 0.244 & 1.060 & 1.748 & 1.353 & 0.799 & 0.662 & 0.653 & 0.655 \\
\glsentryshort{fcd} (Noise-init) & 26.16 & 19.41 & 16.01 & 12.61 & 4.427 & 2.596 & 1.895 & 1.714 \\
\bottomrule
\end{tabular}
}
\caption{\footnotesize MOSES wall-clock generation results on an NVIDIA H100, batch size 128.}
\label{tab:wallclock-runtime}
\end{table}

\paragraph{Sensitivity to the number of transport edits.}
We vary $N$ using a fixed contrastively trained MOSES checkpoint, categorical-noise
initialization, and 5,000 samples per setting with 200- or 500-step chains.
\Cref{tab:transport-edit-sensitivity} shows similar \gls{vun} for $N=1$, $5$, and $10$,
while $N=100$ substantially reduces validity and \gls{vun} scores.
\begin{table}[H]
\centering
\footnotesize
\begin{tabular}{ccccccc}
\toprule
& \multicolumn{3}{c}{200 steps} & \multicolumn{3}{c}{500 steps} \\
\cmidrule(lr){2-4}\cmidrule(lr){5-7}
$N$ & Validity $\uparrow$ & Novelty $\uparrow$ & \acrshort{vun} $\uparrow$ & Validity $\uparrow$ & Novelty $\uparrow$ & \acrshort{vun} $\uparrow$ \\
\midrule
1 & 0.7566 & 0.9831 & 0.7438 & 0.8896 & 0.9676 & 0.8605 \\
5 & 0.7664 & 0.9838 & 0.7540 & 0.8891 & 0.9635 & 0.8560 \\
10 & 0.7582 & 0.9855 & 0.7472 & 0.8836 & 0.9631 & 0.8505 \\
100 & 0.4592 & 0.9993 & 0.2862 & 0.5928 & 0.9995 & 0.4078 \\
\bottomrule
\end{tabular}
\caption{\footnotesize \textbf{Sensitivity to the number of transport edits.}
Noise-initialized MOSES generation. All metrics are fractions.}
\label{tab:transport-edit-sensitivity}
\end{table}

\paragraph{Proposal-ranking cost.}
\Cref{tab:proposal-ranking-cost} compares exact energy-difference ranking with
gradient scoring for single-edge alternatives.
Exact ranking is faster at 16 nodes, but approximately $10\times$ and $159\times$
slower at 32 and 64 nodes, respectively.
Exact energy differences are still used in the \gls{mh} acceptance step.
\begin{table}[H]
\centering
\footnotesize
\begin{tabular}{rrrr}
\toprule
Nodes & Edge candidates & Exact ranking (ms) & Gradient scoring (ms) \\
\midrule
16 & 120 & 18.3 & 29.9 \\
32 & 496 & 286.7 & 28.8 \\
64 & 2016 & 4658.2 & 29.3 \\
\bottomrule
\end{tabular}
\caption{\footnotesize \textbf{Proposal-ranking time.} Single-edge candidates only.}
\label{tab:proposal-ranking-cost}
\end{table}

\paragraph{Graph-size scaling.}
\Cref{tab:graph-size-scaling} reports wall-clock time and peak memory per network
evaluation on one NVIDIA H100 with batch size 1.
\gls{gem}'s forward costs are comparable to the baselines, while gradient
computation includes an additional backward pass and increases runtime and memory use.
The dense edge representation is the main memory bottleneck.
\begin{table}[H]
\centering
\footnotesize
\setlength{\tabcolsep}{5pt}
\begin{tabular}{rcccc}
\toprule
Nodes & \glsentryshort{gem} forward & \glsentryshort{gem} gradient & DiGress forward & DeFoG forward \\
\midrule
64 & 13.2 / 0.09 & 29.3 / 0.31 & 12.5 / 0.09 & 15.2 / 0.12 \\
128 & 14.1 / 0.16 & 30.1 / 0.95 & 13.8 / 0.16 & 16.2 / 0.19 \\
256 & 26.2 / 0.44 & 51.0 / 3.51 & 26.0 / 0.44 & 31.6 / 0.47 \\
\bottomrule
\end{tabular}
\caption{\footnotesize \textbf{Runtime and memory scaling.}
Time (ms) / peak memory (GiB) per network evaluation, on an NVIDIA H100 with batch size 1.
\glsentryshort{gem} gradient includes the forward and backward passes.}
\label{tab:graph-size-scaling}
\end{table}

\subsection{Geodesic analysis: validity along geodesics}
We compare GEM's energy-weighted geodesic to a cost-only baseline along interpolated paths between
matched MOSES molecules. We align permutations using the node-matching $c_{\mathrm{FGW}}$ cost
(\Cref{sec:gem}, \eqref{eq:fgw-cost}; node-matching details in \Cref{app:matching}),
and then use the aligned local cost $c_{\mathrm{loc}}$ \eqref{eq:loc-cost} to evaluate distances,
which is equivalent under the fixed alignment. 

\paragraph{Geodesics.}
Paths are parameterized as cubic B-splines in the
node/edge probability simplex with fixed endpoints and learned interior control points (8 control
points, degree 3). Optimization minimizes the energy-weighted path length $L_\theta$ from
\Cref{app:geodesic-details} (with the $\exp(\beta V_\theta)$ weighting) plus an average-energy
regularizer. We use $\beta=0.1$, length weight $\lambda_L=0.1$, energy weight $\lambda_{ME}=1.0$,
2000 Adam iterations, and a learning rate of $10^{-3}$ \citep{kingma2014adam}.

Validity is estimated by sampling 16 arc-length--uniform locations per path and drawing 16
discrete graphs per location. Validity is the fraction of RDKit-valid and connected samples, averaged
along each path and then across pairs within each distance bin. We use three distance bins:
$[5,10)$, $[10,15)$, and $[15,20)$, with 256 molecules per bin.

\subsection{Energy diagnostics}
\label{app:energy-diagnostics}
We evaluate a MOSES checkpoint trained with contrastive learning on independent
random samples of 10,000 training and 10,000 held-out test molecules.
\Cref{tab:energy-diagnostics} compares their energies with uniform categorical
noise. Held-out molecules receive similar mean energies to training molecules,
while noise has substantially higher energy.

\begin{table}[H]
\centering
\footnotesize
\begin{tabular}{lc}
\toprule
Population & Energy (mean $\pm$ std.) \\
\midrule
MOSES train & $-274.26\pm55.44$ \\
MOSES test & $-273.88\pm55.35$ \\
Uniform categorical noise & $253.80\pm58.12$ \\
\bottomrule
\end{tabular}
\caption{\footnotesize \textbf{Energy diagnostics.}
Mean and standard deviation across samples under the same contrastively trained
MOSES checkpoint.}
\label{tab:energy-diagnostics}
\end{table}

We also run 1,024 data-initialized chains for 300 temperature-annealed mixing
steps and save ten stages. Across these stages, mean energy and validity have
correlation $r=-0.94$. Across sampler settings in a sweep using a contrastively
trained checkpoint, terminal energy and \gls{vun} have correlation $r=-0.72$.
These correlations describe aggregate relationships across stages and sampler
settings.

\section{Noise sensitivity of property regressors}
\label{app:regressor-noise}
\paragraph{Property regressor training.}
Let $(x_1,\zeta)$ denote a molecule (graph) and its scalar property ($\zeta$). A clean regressor $f_\phi$ is
trained on valid molecules by minimizing
\begin{equation}
\min_\phi~\mathbb{E}_{(x_1,\zeta)\sim \mathcal{D}}
\big[\|f_\phi(x_1)-\zeta\|_2^2\big].
\end{equation}
For noise-conditioned regressors, we sample a time $t\sim p(t)$ and a noisy graph
$x_t\sim q_t(x_t\mid x_1)$ using the same discrete noising kernel as the forward process
(marginal or uniform). The regressor takes $(x_t,t)$ as input (time is fed as an additional scalar
feature) and is trained with
\begin{equation}
\min_\phi~\mathbb{E}_{(x_1,\zeta)\sim \mathcal{D},~t\sim p(t),~x_t\sim q_t(\cdot\mid x_1)}
\big[\|f_\phi(x_t,t)-\zeta\|_2^2\big].
\end{equation}
During \gls{gem} sampling we do not explicitly track diffusion time, so we use an energy-based proxy to
set $t$: we linearly increase $t$ between the noise and data energy bands and clamp to $t=1$ once
the chain reaches the data distribution, using $t=1$ for the remainder of generation.

\begin{figure}[t]
\centering
\begin{tikzpicture}
\begin{axis}[
  width=0.70\linewidth,
  height=0.42\linewidth,
  xlabel={Denoising time $t$},
  ylabel={Test \acrshort{mae}},
  xmin=0, xmax=1,
  ymin=0, ymax=1.1,
  grid=both,
  grid style={line width=0.3pt, draw=gray!30},
  tick label style={font=\small},
  label style={font=\small},
  legend style={font=\fontsize{6}{7}\selectfont, draw=none, fill=none, cells={anchor=west}, at={(1.01,1.0)}, anchor=north east},
  legend columns=2,
  legend image post style={xscale=0.5},
]
\addplot+[thick, blue, solid, mark=none] coordinates {
  (0.0, 0.25916087890625)
  (0.2, 0.257115021484375)
  (0.4, 0.252433107421875)
  (0.6, 0.235796013671875)
  (0.8, 0.1833322119140625)
  (1.0, 0.02114986657714844)
};
\addlegendentry{\acrshort{tpsa}/50 (margi.)}

\addplot+[thick, blue, dashed, mark=none] coordinates {
  (0.0, 0.25931730078125)
  (0.2, 0.257358630859375)
  (0.4, 0.252378265625)
  (0.6, 0.240960873046875)
  (0.8, 0.20287095703125)
  (1.0, 0.024920642578125)
};
\addlegendentry{\acrshort{tpsa}/50 (unif.)}

\addplot+[thick, orange!80!black, solid, mark=none] coordinates {
  (0.0, 0.4423901184)
  (0.2, 0.4363820099)
  (0.4, 0.4203746033)
  (0.6, 0.3917442139)
  (0.8, 0.3152096558)
  (1.0, 0.0594828186)
};
\addlegendentry{logS (margi.)}

\addplot+[thick, orange!80!black, dashed, mark=none] coordinates {
  (0.0, 0.4384821198)
  (0.2, 0.4365656372)
  (0.4, 0.4292946228)
  (0.6, 0.4080485626)
  (0.8, 0.3465020111)
  (1.0, 0.0561543186)
};
\addlegendentry{logS (unif.)}

\addplot+[thick, purple!70!black, solid, mark=none] coordinates {
  (0.0, 0.7199128723)
  (0.2, 0.7027059631)
  (0.4, 0.6730399292)
  (0.6, 0.6084385590)
  (0.8, 0.4690968353)
  (1.0, 0.0782309380)
};
\addlegendentry{logP (margi.)}

\addplot+[thick, purple!70!black, dashed, mark=none] coordinates {
  (0.0, 0.7122465149)
  (0.2, 0.7079793396)
  (0.4, 0.6892063049)
  (0.6, 0.6472913635)
  (0.8, 0.5287614197)
  (1.0, 0.0743568142)
};
\addlegendentry{logP (unif.)}

\addplot+[thick, green!60!black, solid, mark=none] coordinates {
  (0.0, 0.0852654148)
  (0.2, 0.0842592201)
  (0.4, 0.0810060917)
  (0.6, 0.0746299919)
  (0.8, 0.0599623322)
  (1.0, 0.0234130266)
};
\addlegendentry{\acrshort{qed} (margi.)}

\addplot+[thick, green!60!black, dashed, mark=none] coordinates {
  (0.0, 0.0857948933)
  (0.2, 0.0853343445)
  (0.4, 0.0841610649)
  (0.6, 0.0802927971)
  (0.8, 0.0680123455)
  (1.0, 0.0258289015)
};
\addlegendentry{\acrshort{qed} (unif.)}
\end{axis}
\end{tikzpicture}
\caption{\footnotesize Test \acrshort{mae} of time/noise-conditioned property regressors versus denoising time $t$
(0 = fully noised, 1 = clean/denoised). Each point averages 10k test molecules. Solid lines are
marginal-noise regressors; dashed lines are uniform-noise regressors. \acrshort{tpsa} \acrshort{mae} is scaled by
$1/50$ for visualization. Errors are large for noisy inputs and only become reasonable around
$t=1$.}
\label{fig:regressor-noise}
\end{figure}

Property regressors are trained on clean molecules, so evaluating them on noisy graphs is
ill-posed when bonds and valences are corrupted. We measure the degradation by computing test
\gls{mae} as a function of denoising time $t$ along the diffusion path. We use 10k test molecules and
report four representative properties (\gls{tpsa}, logS, logP, \gls{qed}). As expected, errors are large for
noisy inputs and only become reasonable at $t=1$ (fully denoised, clean samples).
Figure~\ref{fig:regressor-noise} summarizes the trend. Solid lines are marginal-noise regressors;
dashed lines are uniform-noise regressors.

\begin{table}[t]
\centering
\small
\begingroup
\setlength{\tabcolsep}{4pt}
\begin{tabular}{lccc}
\toprule
Regressor & logS $\ge$ -2.25 & \acrshort{qed} $\ge$ 0.9 & logP $\le$ 1.5 \\
 & \glsentryshort{cvun} $\uparrow$ & \glsentryshort{cvun} $\uparrow$ & \glsentryshort{cvun} $\uparrow$ \\
\midrule
Clean & 0.531 & 0.154 & 0.552 \\
Noise-conditioned  & \textbf{0.836} & \textbf{0.422} & \textbf{0.846} \\
\bottomrule
\end{tabular}
\endgroup
\caption{\footnotesize \textbf{Regressor ablations (\gls{gem}).} Clean regressors are time-independent and trained only on valid molecules. Noise-conditioned regressors are time-dependent and trained on uniformly noised graphs. We report \gls{cvun} for each property condition on MOSES.}
\label{tab:regressor-ablations}
\end{table}

\section{Geodesic analysis details}
\label{app:geodesic-details}

\paragraph{Path optimization.}
We represent a continuous path $\gamma(t)$ in the embedding space and optimize the energy-weighted length
\begin{equation}
L_\theta(\gamma)=\int_0^1 \exp\!\big(\beta V_\theta(\hat{\gamma}(t))\big)\|\dot{\hat{\gamma}}(t)\|_{\mathrm{loc}}\,dt
\end{equation}
with an average-energy regularizer $\lambda_{ME}\int_0^1 V_\theta(\hat{\gamma}(t))\,dt$. The cost-based
baseline uses the same endpoint pairing and interpolates between the coupled graphs without the additional energy-weighting or average-energy regularization.

\paragraph{Spline parameterization.}
We implement geodesic paths using a cubic B-spline over learnable control-point logits for nodes and
edges. The spline lives in the probability simplex (node/edge categorical distributions), and
discrete graphs are sampled only for visualization and metric evaluation. The optimization minimizes
the relative length ratio $L_{\mathrm{spline}}/L_{\mathrm{linear}} - 1$, where the segment lengths are
weighted by $\exp\!\big(\beta V_\theta(x_\tau)\big)$ along the path.

\paragraph{Discrete evaluation.}
We discretize each continuous path at evenly spaced points in normalized arc length. At each point
we sample categorical graphs from the path distribution to compute validity and energy statistics.
We compute average validity along each path and plot it against the endpoint distance induced by the
\gls{fgw} node-matching cost $c_{\mathrm{FGW}}$.
Validity estimation and binning details are in \Cref{app:exp-setup}.

\section{Contrastive loss objective derivation}\label{app:energy-refinement}
Given the model density (where $\beta_{\mathrm{mh}}$ is a constant):
\begin{equation}
p_\theta(x)=\frac{e^{-\beta_{\mathrm{mh}} V_\theta(x)}}{Z_\theta},
\qquad
Z_\theta=\sum_x e^{-\beta_{\mathrm{mh}} V_\theta(x)}.
\end{equation}
let us consider maximum likelihood estimation:
\begin{equation}
\theta^\star=\arg\max_\theta\;\mathbb{E}_{x\sim p_{\text{data}}}\big[\log p_\theta(x)\big]
\;\;\Longleftrightarrow\;\;
\theta^\star=\arg\min_\theta\;\mathcal{J}(\theta),
\end{equation}
with
\begin{equation}
\mathcal{J}(\theta):=-\mathbb{E}_{p_{\text{data}}}\big[\log p_\theta(x)\big]
=\mathbb{E}_{p_{\text{data}}}\big[\beta_{\mathrm{mh}} V_\theta(x)\big]+\log Z_\theta.
\end{equation}
Taking gradients gives
\begin{equation}
\nabla_\theta \mathcal{J}(\theta)
=\mathbb{E}_{p_{\text{data}}}\big[\beta_{\mathrm{mh}}\nabla_\theta V_\theta(x)\big]
+\nabla_\theta \log Z_\theta,
\end{equation}
where
\begin{equation}
\begin{aligned}
\nabla_\theta \log Z_\theta
&=\frac{1}{Z_\theta}\nabla_\theta Z_\theta
=\frac{1}{Z_\theta}\sum_x e^{-\beta_{\mathrm{mh}} V_\theta(x)}\big(-\beta_{\mathrm{mh}}\nabla_\theta V_\theta(x)\big)\\
&=-\mathbb{E}_{p_\theta}\big[\beta_{\mathrm{mh}}\nabla_\theta V_\theta(x)\big].
\end{aligned}
\end{equation}
Therefore, the gradient simplifies to
\begin{equation}
\boxed{
\nabla_\theta \mathcal{J}(\theta)
=\beta_{\mathrm{mh}}\left(\mathbb{E}_{p_{\text{data}}}\big[\nabla_\theta V_\theta(x)\big]
-\mathbb{E}_{p_\theta}\big[\nabla_\theta V_\theta(x)\big]\right)
}.
\end{equation}
This motivates the contrastive loss objective \(\mathcal{L}_{\mathrm{CL}}\) introduced in \eqref{eq:cl}, where we approximate the expectation over \(p_\theta\) using samples initialized either at uniform noise and data samples in 50\%/50\% proportions for general training, or exclusively at data samples (0\%/100\%) for the specialized training of data-initialized unconditional generation models. We then iteratively run a sequence of \((\text{greedy} + N_{\mathrm{CL}})\) proposals, explicitly detaching gradients flowing through the sampler transitions (i.e., Markov chain proposal and acceptance steps) to avoid backpropagation through the stochastic sampling process itself, ensuring stable estimation.

\section{Broader Impacts}
\label{app:broader-impacts}

GEM is a methodological contribution for generative modeling of discrete structured data, evaluated here on public graph benchmarks. Potential positive impacts include more efficient exploration of molecular design spaces, improved incorporation of structural or property constraints, and better tools for scientific discovery workflows. Potential negative impacts arise from the same generative capabilities: molecular generation methods could be misused to propose unsafe, toxic, or otherwise undesirable compounds if deployed without domain-specific screening and expert oversight. Our experiments do not validate generated molecules for synthesis, biological activity, safety, or deployability. Responsible use should therefore pair such methods with established cheminformatics filters, toxicity and synthesizability checks, domain-expert review, and application-specific governance before any real-world use.

\section{Additional molecular results}
\label{app:published-moses}

\Cref{tab:published-moses} compares additional MOSES baselines with \gls{gem}.
We report these baselines separately from those in \Cref{tab:uncond-moses}
due to differences in backbones and sampling methods.
We recompute DiGress \gls{vun}
and \gls{fcd} from released samples and derive Cometh's \gls{vun} from reported
validity, uniqueness, and novelty, using its predictor--corrector (PC) variant.
Both baselines achieve lower \gls{fcd} than our main \gls{gem}
configuration, but with substantially lower \gls{vun}
(DiGress: $0.812$ vs.\ $0.923$).

\begin{table}[H]
\centering
\footnotesize
\begin{tabular}{lcc}
\toprule
Method & \acrshort{vun} $\uparrow$ & \acrshort{fcd} $\downarrow$ \\
\midrule
DiGress~\citep{vignac2022digress} & 0.812 & 1.19 \\
Cometh (PC)~\citep{siraudin2024cometh} & 0.837 & 1.27 \\
\gls{gem} & 0.923 & 1.39 \\
\bottomrule
\end{tabular}
\caption{\footnotesize \textbf{Additional molecular baselines on MOSES.}
Noise-initialized generation. Higher \acrshort{vun} and lower \acrshort{fcd} are better.}
\label{tab:published-moses}
\end{table}

\paragraph{Individual QM9 metrics.}
\Cref{tab:qm9-components} reports individual QM9 metrics from the VFM and DeFoG papers
and our \gls{gem} evaluation, with VFM's \gls{vun} computed from its reported validity, uniqueness, and novelty.
Higher novelty drives \gls{gem}'s higher \gls{vun}, while VFM and DeFoG achieve lower \gls{fcd}.

\begin{table}[H]
\centering
\footnotesize
\begin{tabular}{lccccc}
\toprule
Method & Validity $\uparrow$ & Uniqueness $\uparrow$ & Novelty $\uparrow$ & \acrshort{vun} $\uparrow$ & \acrshort{fcd} $\downarrow$ \\
\midrule
\gls{vfm}~\citep{eijkelboom2024vfm} & 0.998 & 1.000 & 0.490 & 0.489 & 0.44 \\
DeFoG~\citep{qinmadeira2024defog} & 0.993 & 0.963 & 0.364 & 0.348 & 0.81 \\
\gls{gem} & 0.987 & 0.991 & 0.626 & 0.612 & 0.92 \\
\bottomrule
\end{tabular}
\caption{\footnotesize \textbf{Individual QM9 metrics.} Noise-initialized generation.}
\label{tab:qm9-components}
\end{table}

\section{Non-molecular graph generation}
\label{app:nonmolecular}

We evaluate \gls{gem} on Planar, comprising unlabeled 64-node graphs, and stochastic block model (SBM) graphs with up to 187 nodes.
For Planar, we evaluate \gls{gem} and released DiGress samples using the native DiGress implementation for structural metrics and graph-isomorphism novelty~\citep{vignac2022digress}.
Both methods use 1,000 sampling steps.
For SBM, we follow the DeFoG evaluation protocol and take baseline results from Table~7 of \citet{qinmadeira2024defog}.
Our SBM model uses the DeFoG backbone (17.5M parameters) with a permutation-invariant energy head (0.1M parameters), 1,000 sampling steps, and transport proposals with $N=10$.

\begin{table}[H]
\centering
\footnotesize
\setlength{\tabcolsep}{3pt}
\begin{tabular}{@{}lccccc@{}}
\toprule
Method & Degree $\downarrow$ & Clustering $\downarrow$ & Orbit $\downarrow$ & Spectral $\downarrow$ & \acrshort{vun} $\uparrow$ \\
\midrule
\multicolumn{6}{l}{Planar (64 nodes)} \\
DiGress & 0.000622 & 0.056330 & 0.009802 & \textbf{0.006212} & \textbf{0.8500} \\
\gls{gem} & \textbf{0.000375} & \textbf{0.051806} & \textbf{0.005711} & 0.007292 & 0.8075 \\
\midrule
\multicolumn{6}{l}{SBM (up to 187 nodes)} \\
DiGress & 0.0018 & 0.0485 & 0.0415 & 0.0045 & 0.600 \\
DeFoG & $\mathbf{0.0006}\pm0.0023$ & $0.0517\pm0.0012$ & $0.0556\pm0.0739$ & $0.0054\pm0.0012$ & $\mathbf{0.900}\pm0.051$ \\
CatFlow (VFM) & 0.0012 & 0.0498 & 0.0357 & -- & 0.850 \\
Cometh & $0.0020\pm0.0003$ & $0.0498\pm0.0000$ & $0.0383\pm0.0051$ & $\mathbf{0.0024}\pm0.0003$ & $0.750\pm0.037$ \\
\gls{gem} & $0.0012\pm0.0005$ & $\mathbf{0.0457}\pm0.0008$ & $\mathbf{0.0343}\pm0.0033$ & $0.0048\pm0.0005$ & $\mathbf{0.900}\pm0.018$ \\
\bottomrule
\end{tabular}
\caption{\footnotesize \textbf{Non-molecular graph generation.}
The first four metrics are maximum mean discrepancies (MMD).
SBM \gls{gem} results are means $\pm$ standard deviations over four seeds.
Reported baseline uncertainties are retained, and a dash denotes an unavailable value.
Best values within each benchmark are bold, including ties in reported means.}
\label{tab:nonmolecular}
\end{table}

On Planar, \gls{gem} improves degree, clustering, and orbit MMD and validity ($0.95$ vs.\ $0.85$), while DiGress achieves better spectral MMD, higher novelty ($1.00$ vs.\ $0.85$), and higher \gls{vun} scores.
On SBM, \gls{gem} achieves the lowest mean clustering and orbit MMD among the listed methods and ties the best mean \gls{vun} of $0.90$.
Overall, these results show that \gls{gem} remains competitive on these non-molecular graph benchmarks while providing an explicit interface to the learned unnormalized density landscape through its scalar energy.

\newpage
\section*{NeurIPS Paper Checklist}

\begin{enumerate}

\item {\bf Claims}
    \item[] Question: Do the main claims made in the abstract and introduction accurately reflect the paper's contributions and scope?
    \item[] Answer: \answerYes{}.
    \item[] Justification: Our claims are restricted to the graph domain. The empirical evaluation focuses primarily on molecular datasets, with additional results on the non-molecular Planar and SBM datasets. We discuss the limitations and tradeoffs of the chosen representation.

\item {\bf Limitations}
    \item[] Question: Does the paper discuss the limitations of the work performed by the authors?
    \item[] Answer: \answerYes{}.
    \item[] Justification: The discussion notes the main computational trade-off of energy-gradient sampling, while the appendix states the approximation and guarantee scope and reports robustness, sensitivity, and wall-clock measurements.

\item {\bf Theory assumptions and proofs}
    \item[] Question: For each theoretical result, does the paper provide the full set of assumptions and a complete (and correct) proof?
    \item[] Answer: \answerYes{}.
    \item[] Justification: The paper states the assumptions and approximation scope behind the proposal derivations, and the appendix specifies that formal stationarity guarantees apply only to the MH mixing stage.

    \item {\bf Experimental result reproducibility}
    \item[] Question: Does the paper fully disclose all the information needed to reproduce the main experimental results of the paper to the extent that it affects the main claims and/or conclusions of the paper (regardless of whether the code and data are provided or not)?
    \item[] Answer: \answerYes{}.
    \item[] Justification: The paper describes datasets, metrics, algorithms, architecture, hyperparameters, sample counts, baseline handling, and robustness/runtime details; code is provided.

\item {\bf Open access to data and code}
    \item[] Question: Does the paper provide open access to the data and code, with sufficient instructions to faithfully reproduce the main experimental results, as described in supplemental material?
    \item[] Answer: \answerYes{}.
    \item[] Justification: Code is provided to reproduce the paper's main claims.

\item {\bf Experimental setting/details}
    \item[] Question: Does the paper specify all the training and test details (e.g., data splits, hyperparameters, how they were chosen, type of optimizer) necessary to understand the results?
    \item[] Answer: \answerYes{}.
    \item[] Justification: Training schedules, optimizers, hyperparameters, sampling settings, architectures, sample counts, and evaluation protocols are specified in the main text and appendix.

\item {\bf Experiment statistical significance}
    \item[] Question: Does the paper report error bars suitably and correctly defined or other appropriate information about the statistical significance of the experiments?
    \item[] Answer: \answerYes{}.
    \item[] Justification: The appendix reports mean and standard deviation for training-seed and sampling-seed variability, and states which sources of randomness are measured.

\item {\bf Experiments compute resources}
    \item[] Question: For each experiment, does the paper provide sufficient information on the computer resources (type of compute workers, memory, time of execution) needed to reproduce the experiments?
    \item[] Answer: \answerYes{}.
    \item[] Justification: The appendix reports hardware for wall-clock experiments and provides training, sampling, and runtime details needed to estimate compute requirements.
    
\item {\bf Code of ethics}
    \item[] Question: Does the research conducted in the paper conform, in every respect, with the NeurIPS Code of Ethics \url{https://neurips.cc/public/EthicsGuidelines}?
    \item[] Answer: \answerYes{}.
    \item[] Justification: The work uses public molecular benchmarks and standard computational experiments, and does not involve human subjects or private data.

\item {\bf Broader impacts}
    \item[] Question: Does the paper discuss both potential positive societal impacts and negative societal impacts of the work performed?
    \item[] Answer: \answerYes{}.
    \item[] Justification: Potential positive scientific uses and dual-use risks for molecular generation are discussed in \Cref{app:broader-impacts}.
    
\item {\bf Safeguards}
    \item[] Question: Does the paper describe safeguards that have been put in place for responsible release of data or models that have a high risk for misuse (e.g., pre-trained language models, image generators, or scraped datasets)?
    \item[] Answer: \answerNA{}.
    \item[] Justification: The paper does not release high-risk pretrained models or scraped datasets requiring special safeguards.

\item {\bf Licenses for existing assets}
    \item[] Question: Are the creators or original owners of assets (e.g., code, data, models), used in the paper, properly credited and are the license and terms of use explicitly mentioned and properly respected?
    \item[] Answer: \answerYes{}.
    \item[] Justification: Existing datasets, software, and baselines are cited in the paper; released code will include the relevant license information.

\item {\bf New assets}
    \item[] Question: Are new assets introduced in the paper well documented and is the documentation provided alongside the assets?
    \item[] Answer: \answerYes{}.
    \item[] Justification: The provided code documents how to reproduce the method.

\item {\bf Crowdsourcing and research with human subjects}
    \item[] Question: For crowdsourcing experiments and research with human subjects, does the paper include the full text of instructions given to participants and screenshots, if applicable, as well as details about compensation (if any)? 
    \item[] Answer: \answerNA{}.
    \item[] Justification: The paper does not involve crowdsourcing or human-subject experiments.

\item {\bf Institutional review board (IRB) approvals or equivalent for research with human subjects}
    \item[] Question: Does the paper describe potential risks incurred by study participants, whether such risks were disclosed to the subjects, and whether Institutional Review Board (IRB) approvals (or an equivalent approval/review based on the requirements of your country or institution) were obtained?
    \item[] Answer: \answerNA{}.
    \item[] Justification: The paper does not involve crowdsourcing or human-subject experiments.

\item {\bf Declaration of LLM usage}
    \item[] Question: Does the paper describe the usage of LLMs if it is an important, original, or non-standard component of the core methods in this research? Note that if the LLM is used only for writing, editing, or formatting purposes and does \emph{not} impact the core methodology, scientific rigor, or originality of the research, declaration is not required.
    \item[] Answer: \answerNA{}.
    \item[] Justification: The core method does not use LLMs as an important, original, or non-standard component.

\end{enumerate}

\end{document}